\documentclass[11pt]{article}
    
    \usepackage[preprint]{acl}
    
    \usepackage{times}
    \usepackage{latexsym}
    \usepackage{amsmath}
    \usepackage{amsfonts}
    \usepackage{booktabs}
    \usepackage{inconsolata}
    \usepackage{tabularx}
    \newcolumntype{Y}{>{\raggedright\arraybackslash}X}
    
    \usepackage[T1]{fontenc}
    
    \usepackage[utf8]{inputenc}
    
    \usepackage{microtype}

    \usepackage{graphicx}
    \usepackage{xcolor}

    \title{The Interlingua Hypothesis: LLMs Translate\\via a Latent Task-agnostic Feature Space}

    \author{
  \textbf{Jacob Brinton$^1$}\ \ \ \ \ 
  \textbf{Jannik Brinkmann$^2$}\ \ \ \ \ 
  \textbf{Mark Crovella$^1$}\ \ \ \ \ 
  \textbf{Aaron Mueller$^1$}
\\
  \textsuperscript{1}Boston University\ \ \ \ \ \ \  
  \textsuperscript{2}Technische Universität Clausthal
\\
  \texttt{jbrin@bu.edu},\ \ \ \texttt{jannik.brinkmann@uni-mannheim.de},\\ \texttt{crovella@bu.edu},\ \ \ \texttt{amueller@bu.edu}
}

\begin{document}
    \maketitle
    \begin{abstract}
    Large language models (LLMs) have recently demonstrated improved machine translation performance over strong supervised baselines. This raises questions as to what mechanisms underlie how LLMs perform machine translation between languages. Motivated by recent interpretability findings---namely, that LLMs use massively multilingual latent feature representations to perform language modeling---we propose the interlingua hypothesis. The hypothesis holds that language models translate by reading a source sentence into a latent feature space, and generate a target sentence by reading from the latent feature space. We show three lines of evidence in support of this hypothesis: (1) variance in BLEU across language pairs is largely predictable from language-specific competences with no language pair--specific interaction terms; (2) many model components are causally influential in both monolingual tasks and translation tasks; and (3) fine-tuning on monolingual data recovers a large proportion of translation improvements relative to fine-tuning on aligned documents. Together, these provide convergent evidence in support of the interlingua hypothesis, and suggest new ways of understanding and improving how LLMs can be leveraged to perform translation tasks.
    \end{abstract}

    \section{Introduction}
    
    Machine translation has long been an influential area of natural language processing. The attention mechanism was first proposed to improve machine translation performance \citep{bahdanau2015neuralmachinetranslationjointly,luong-etal-2015-effective,vaswani2017attention}; one consequence of attention was the emergence of language models that could be efficiently trained on massive corpora \citep{devlin-etal-2019-bert,radford2018improving}. Recently, large language models (LLMs) have revolutionized many areas of natural language processing, but have been relatively slowly adopted in machine translation (MT).
    
    A reason for the slow adoption of LLMs in MT has been a focus on low-resource languages, where LLMs face significant challenges \citep{hendy2023goodgptmodelsmachine,robinson2023chatgptmtcompetitivehigh}. Because LLMs require large training corpora, they are difficult to train effectively on low-resource languages. However, they also hold significant promise: even when not trained on a given language, LLMs can be prompted to translate from or into a language they have seen very little of in their training data \citep{cahyawijaya-etal-2024-llms}---e.g., by prompting with a grammar book \citep{tanzer2024a}. Moreover, their representations of high-resource languages could enable cross-lingual transfer \citep{conneau-etal-2020-unsupervised}.
    
    Whether LLMs can deliver on this promise depends in part on the mechanisms underlying how they translate. Specifically, do LLMs rely on task- and language-agnostic mechanisms to translate, or do they rely on more specialized translation mechanisms or language pair--specific mechanisms? If the former, this suggests clear paths toward improving MT performance, potentially without large parallel corpora. Much recent work provides mechanistic evidence supporting the existence of massively multilingual representations in language modeling contexts \citep{wendler-etal-2024-llamas,brinkmann-etal-2025-large,wu2025the}. 
    
    We therefore hypothesize that LLMs perform MT in large part by reusing computational mechanisms for monolingual language modeling. The ``stages of inference'' hypothesis \citep{lad2024the} holds that LLMs devote the first half of their layers to reading an input and composing increasingly abstract concept representations. Then, the latter half of the model reads these concept representations, and uses them to decide which token should be predicted given prior context. If language models perform translation this way (what we call the \textbf{interlingua hypothesis}), then translation would not necessarily require language pair--specific mechanisms; instead, it only requires that a model be capable of reading abstract features from the source language, and generating text for the target language conditioned on those features.
    
    If this hypothesis is true, it would entail the following predictions: (1) Given language pair $(S,T)$, translation performance should be predictable from monolingual capabilities in $S$ and $T$ in isolation without cross-linguistic interaction terms. (2) There should exist task-agnostic representations that are causally relevant for predicting correct outputs in both machine translation \emph{and} monolingual task settings. Finally, (3) adding translation capabilities for a new language should primarily require improvements to monolingual capabilities for that language---i.e., fine-tuning on monolingual corpora should recover a substantial proportion of the improvement in translation performance as fine-tuning on parallel corpora, assuming the model can already effectively handle the other language in the language pair.\footnote{Note that our claims relate to the mechanisms a \emph{fully trained} LLM uses to perform translation. We do not make claims regarding what is contained in the pretraining data; it is possible that parallel pretraining data is necessary for multilingual representations to be learned.}
    
    We investigate each of these three implications, and find positive evidence for each. While no single experiment definitively confirms the interlingua hypothesis, each provides a different type of evidence in support of it. These findings could provide a preliminary explanation for the value of exposure to (largely monolingual) documents in producing models more effective at translation.

    \section{Related Work}
    \paragraph{Massively multilingual feature representations.} Our work is partially motivated by the observation that grammatical concept representations are highly multilingual in large language models, even across typologically distinct languages \citep{brinkmann-etal-2025-large}. Similar evidence has been observed in \citet{wendler-etal-2024-llamas,dumas-etal-2025-separating}. Notably, intervening on grammatical concept representations has predictable effects on model behavior in both monolingual and machine translation settings \citep{brinkmann-etal-2025-large}. These findings do not in themselves confirm the interlingua hypothesis, but they do suggest the existence of abstract feature representations not tied to particular tasks nor languages.

    \paragraph{Interlingua in multilingual NMT.}
    A line of work in neural machine translation (NMT) seeks to engineer interlingua to improve cross-lingual transfer via architectural and training improvements, largely to encoder-decoder architectures (e.g., XLM-R, \citealp{conneau-etal-2020-unsupervised}; M2M, \citealp{fan-etal-2020-beyond}). Prior approaches introduce explicit shared representations via interlingua layers or networks (\citealp{lu-etal-2018-neural}; \citealp{zhu-etal-2020-language}), or bottlenecks that naturally encourage language-independent representations (\citealp{vazquez-etal-2019-multilingual}; \citealp{mao-etal-2023-variable}). These studies stipulate or engineer an interlingua and validate it behaviorally, whereas our study asks whether an interlingua emerges naturally and is causally relevant to MT performance in a decoder-only LM trained on general language data with no such directly implemented incentives.

Precedents to our mechanistic investigation include \citet{vazquez-etal-2020-systematic}, who investigate whether sentence representations in NMT models with attention bottlenecks are language-independent. Others have investigated whether shared or language-specific components are needed at all (\citealp{escolano-etal-2021-multilingual}; \citealp{purason-tattar-2022-multilingual}); these studies investigate by controlling the degree of parameter sharing across languages, and find that fully shared representations underperformed representations with at least some language-specific components. Thus, in smaller-scale settings using primarily parallel data, interlingua typically need to be imposed via the training objective, and this costs some performance. We investigate whether this also holds in contemporary decoder-only models trained on a larger quantity of task- and domain-general language data.
    
    \paragraph{LLMs for machine translation.} Machine translation is difficult when parallel data is limited \citep{koehn-knowles-2017-six}, and large language models do not solve this problem \citep{court-elsner-2024-shortcomings}. Some hope that LLMs could enable cross-lingual transfer. For example, \citet{tanzer2024a} showed that long-context LLMs can translate a previously unseen low-resource language when prompted with a grammar book containing linguistic descriptions and translation examples. However, \citet{aycock2025can} subsequently found that most of the improvement in this setting came from the book’s parallel examples rather than its grammatical explanations, highlighting the importance of parallel data for translation. One of our experiments asks a complementary question: once an LLM has already acquired multilingual representations during pretraining, to what extent can improving its monolingual competence in a low-resource language improve translation without additional parallel data? We provide preliminary evidence for cross-lingual transfer in \S\ref{sec:monolingual-vs-parallel}.
    
    \paragraph{Causal mediation analysis.} Some of our evidence relies on estimates of the causal influence \citep{lewis1973causation} of specific model components. This relies on causal mediation analysis \citep{pearl-2001-cma,vig-etal-2020-cma}, a common technique in the mechanistic interpretability literature \citep[e.g.,][]{finlayson-etal-2021-causal,geiger-etal-2021-causal,heimersheim2024useinterpretactivationpatching,marks2025sparse,mueller-etal-2026-quest}. While the purpose of this study is not to understand the exact functional role of particular model components, we wish to characterize whether the \emph{same} representations influence a model's behavior in multiple task settings and language pairs.
    
    \section{Modeling Translation Performance as a Function of Monolingual Capabilities}\label{sec:linear_v_bilinear}
    
    How much of a model's ability to translate can be explained by its monolingual language modeling capabilities? If a model uses an interlingua to perform translation, then much of its translation capabilities should be explainable as a function of its ability to parse latent features from an input, and produce coherent text in that language. In other words, one should not need language pair--specific terms to explain translation performance, unless a model deploys some separate translation mechanism that does not involve going through an interlingua. To investigate, we define a \textbf{linear model} that predicts translation performance from monolingual competencies.
    
    We use several measures of monolingual competence. First, to measure a model's ability to distinguish grammatical from ungrammatical sentences, we use MultiBLiMP \citep{jumelet-etal-2026-multiblimp}, a multilingual version of the BLiMP \citep{warstadt-etal-2020-blimp-benchmark} benchmark. For a given language, MultiBLiMP accuracy is the fraction of minimal pairs for which the model assigns higher full-sentence log-probability to the grammatical completion over its ungrammatical counterpart. We average this over 200 randomly-selected samples per language. Because Llama and Aya saturate this benchmark for many of the languages in our analysis, we also compute the log-probability \emph{margin}, which is the mean $\log p(\text{grammatical})-\log p(\text{nongrammatical})$ per minimal pair. To overcome issues related to benchmark saturation, we also include accuracy on GlobalMMLU \citep{singh-etal-2025-global}, a multilingual multiple-choice question answering dataset based on MMLU \citep{hendrycks2021measuring}. For each task, we filter the set of languages to those in common between the monolingual dataset and FLORES (Table~\ref{tab:proxies}).\footnote{The following languages are supported across all datasets we use in this section: \texttt{ara, ces, deu, ell, eng, fas, fra, heb, hin, ita, nld, pol, por, ron, rus, spa, tur, ukr}}
    
    For translation, we have language pair $(S, T)$, where $S$ is the source and $T$ is the target language. Translation competence $t_{ST}$ is quantified as a BLEU score for $(S, T)$. To perform translation, we use a 2-shot prompt containing two randomly sampled sentences from FLORES.\footnote{We ensure that the examples do not overlap with the test example.} We use \texttt{sacrebleu} \citep{post-2018-call} to compute BLEUs.
    
    Let $l$ be a monolingual competence estimate using MultiBLiMP or GlobalMMLU. We have $l_S$ and $l_T$ for each language pair. We then learn coefficients $\beta_S$ and $\beta_T$ as well as bias term $\beta_0$ to maximize predictive accuracy on BLEU scores $t_{ST}$ in the following function:
    \begin{equation}
        \beta_S l_S + \beta_Tl_T + \beta_0 = t_{ST}
        \label{eq:linear-model}
    \end{equation}
    
    We also train a \textbf{bilinear model} that is nearly identical, but also contains a multiplicative interaction term:
    \begin{equation}
        \beta_Sl_S + \beta_Tl_T + \beta_{ST}(l_{S}\cdot l_T) + \beta_0 = t_{ST}
    \end{equation}
    If the interlingua hypothesis holds, then the bilinear model should not have significantly greater predictive power than the linear model. This would imply that machine translation performance is better explained by monolingual terms, rather than by the existence of a translation mechanism (which should use terms specific to particular language \emph{pairs}).
    
    \begin{table}[t]
    \centering
    \small
    \setlength{\tabcolsep}{4pt}
    \begin{tabularx}{\linewidth}{@{}lYc@{}}
    \toprule
    Proxy & Measures & $N$ \\
    \midrule
    FLORES perplexity      & fluency / compression ($\downarrow$)   & 24 \\
    MultiBLiMP accuracy    & grammatical acceptability ($\uparrow$) & 18 \\
    MultiBLiMP margin      & grammatical confidence ($\uparrow$)    & 18 \\
    GlobalMMLU accuracy          & world knowledge ($\uparrow$)      & 23 \\
    \bottomrule
    \end{tabularx}
    \caption{Monolingual competence proxies. $N$ is the number of the 24 FLORES languages on which the proxy is defined; $\uparrow$/$\downarrow$ indicates the direction of higher monolingual competence.}
    \label{tab:proxies}
    \end{table}
    
\begin{figure}[t]
    \centering
    \includegraphics[width=\linewidth]{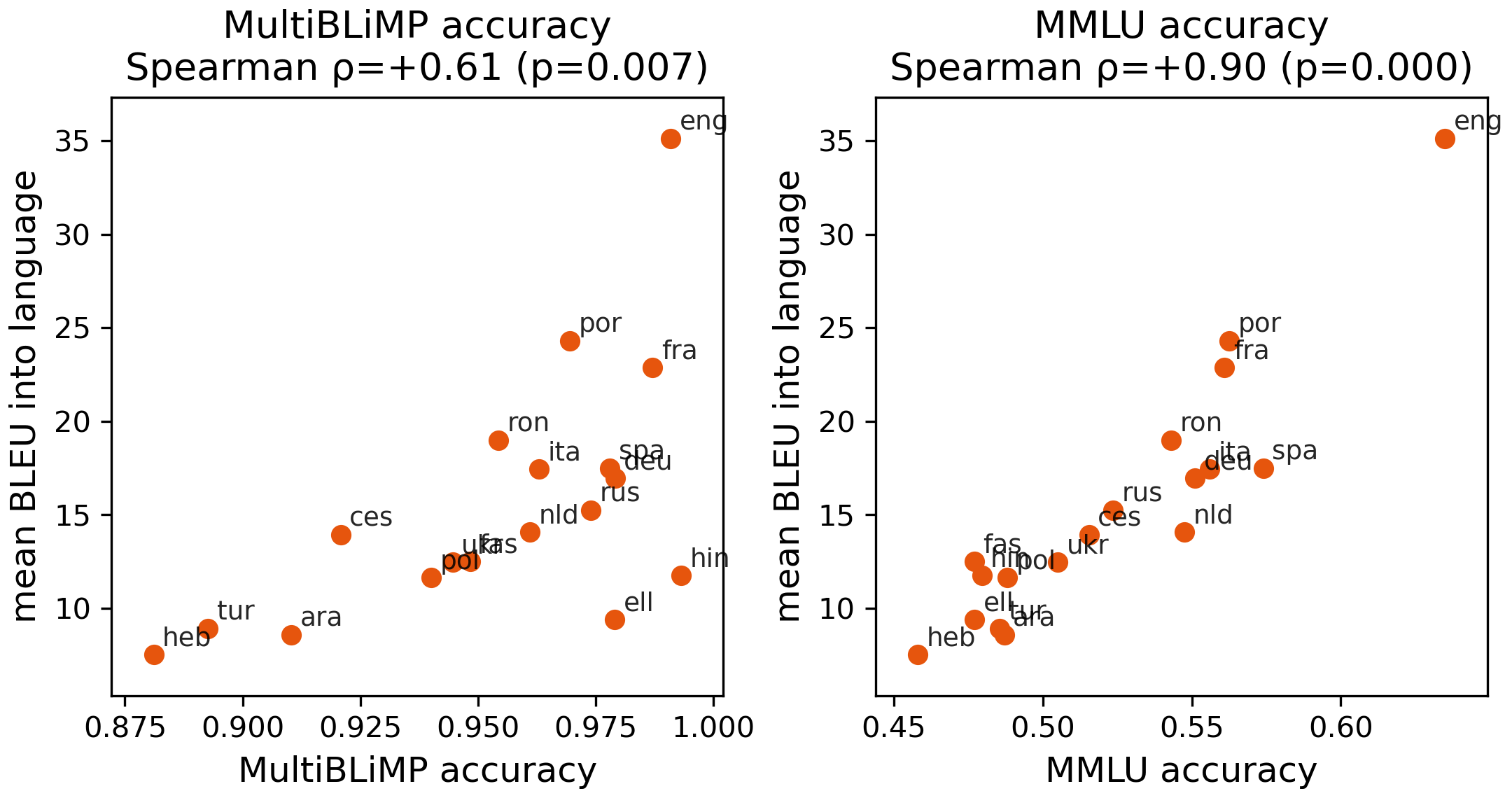}
    \caption{Per-language relationship between two monolingual proxies and the BLEU score for a given target language. Both monolingual terms correlate significantly with BLEU scores, but GlobalMMLU accuracies are a stronger predictor than MultiBLiMP accuracies.}
    \label{fig:proxy}
    \end{figure}

    \paragraph{Monolingual task performance predicts translation competence.}
    Fitting the linear model on the common 18-language subset, monolingual behavioral competence proxies are generally strong predictors of translation performance. MultiBLiMP grammatical accuracy reaches $R^2=0.294$/$0.235$ for Llama/Aya. While significant, this is relatively low; we find that this is largely because MultiBLiMP scores saturate at relatively low BLEU scores, such that it is a good predictor, but only up to middling BLEU scores. In contrast, GlobalMMLU accuracy is a very strong predictor at $R^2=0.739$/$0.510$ for Llama/Aya. Figure~\ref{fig:proxy} plots monolingual competencies against translation performance for each language shared across each evaluation dataset. Qualitatively, the grammatical and MMLU proxies trend closely with translation quality.

    \begin{table}[t]
    \centering
    \resizebox{\linewidth}{!}{
    \setlength{\tabcolsep}{4pt}
    \begin{tabular}{lcccccc}
    \toprule
     & \multicolumn{3}{c}{Llama-3.1-8B} & \multicolumn{3}{c}{Aya-23-8B} \\
    \cmidrule(lr){2-4}\cmidrule(lr){5-7}
    Proxy & $R^2_{\text{lin}}$ & $R^2_{\text{bil}}$ & MAE & $R^2_{\text{lin}}$ & $R^2_{\text{bil}}$ & MAE \\
    \midrule
    Per-token perplexity  & 0.046 & 0.046 & 5.48 & 0.002 & 0.003 & 4.89 \\
    \midrule
    MultiBLiMP accuracy   & 0.294 & 0.294 & 4.50 & 0.235 & 0.236 & 4.22 \\
    MultiBLiMP margin     & 0.324 & 0.324 & 4.82 & 0.233 & 0.234 & 4.27 \\
    GlobalMMLU accuracy         & \textbf{0.739} & 0.741 & \textbf{2.91} & \textbf{0.510} & 0.510 & \textbf{3.56} \\
    \bottomrule
    \end{tabular}}
    \caption{Comparison of the power of monolingual tasks in predicting BLEU scores using the linear ($R^2_{\text{lin}}$) and bilinear ($R^2_{\text{bil}}$) models with several monolingual competence proxies on 18 languages. GlobalMMLU is the strongest predictor. The bilinear interaction term does not add significant predictive power over the linear model.}
    \label{tab:proxy}
    \end{table}

    \paragraph{Adding language pair interaction terms does not increase predictive power.}
    Across each proxy and both models, the bilinear model has virtually the same predictive power as the linear model (Table~\ref{tab:proxy}; see also App.~\ref{app:grids} and App.~\ref{app:interaction}).

    The same trend holds when we predict BLEU scores from the language-specific marginal BLEU scores, computed by averaging BLEU scores across all language pairs for a given source or target language. Decomposing the full BLEU matrix into per-language source and target main effects, $\text{BLEU}_{ST}\approx\mu+\alpha_S+\beta_T$, explains $R^2=0.932$ (Llama) / $0.879$ (Aya) of the centered variance, and a rank-1 multiplicative reconstruction recovers $\approx$$90\%$/$83\%$ of the matrix energy (App.~\ref{app:rank1}). Hence, pairwise translation quality is largely captured by per-language competence, with pairwise interactions adding little predictive power over the simpler model.
    
    \paragraph{Target language competence matters more than source language competence.}
    Source and target competence contribute asymmetrically: changing the target language moves BLEU far more than changing the source. 
    The variance across target languages of mean BLEU exceeds the variance across source languages by $9.3\times$ for Llama and $3.9\times$ for Aya, and the  target coefficient is $1.7$--$5\times$ the source coefficient (MMLU target/source $=2.98$ for Llama, $1.67$ for Aya; MultiBLiMP-margin $4.88$/$2.30$). This may be because it is easier for models to extract meaning from the source language than produce fluent outputs in the target language; alternatively, it may  be an artifact of relying on an n-gram matching metric such as BLEU score.

    \section{Many Translation-relevant Components Also Perform Monolingual Tasks}
    
    Our previous results show that monolingual capabilities are predictors of translation performance, but also that language pair interactions are not strong predictors. This provides correlational evidence in support of our hypothesis, but not causal evidence. To obtain causal evidence, we now perform an analysis based on causal mediation analysis \citep{pearl-2001-cma,vig-etal-2020-cma}.
    
    Past work has found that language models use the first half of their layers to form progressively more abstract representations of the latent features in a given input \citep{lad2024the}. We hypothesize that language models could repurpose this language modeling machinery to perform translation by reading the source language into a latent feature space, and then using the latter half of its layers to read from this feature space and produce the translation in the output language. If this is true, then we should be able to find model representations or components that are causally influential in both machine translation and language modeling settings. To obtain causal evidence in support of this view, we now perform mechanistic experiments by patching model components. We show that the most influential attention heads for producing correct translations also strongly mediate a language model's ability to produce grammatical outputs in monolingual settings.
    
    \subsection{Using GCM to Identify Translation-relevant Components}\label{sec:gcm}
    
    \begin{figure*}[t]
    \centering
    \includegraphics[width=0.8\linewidth]{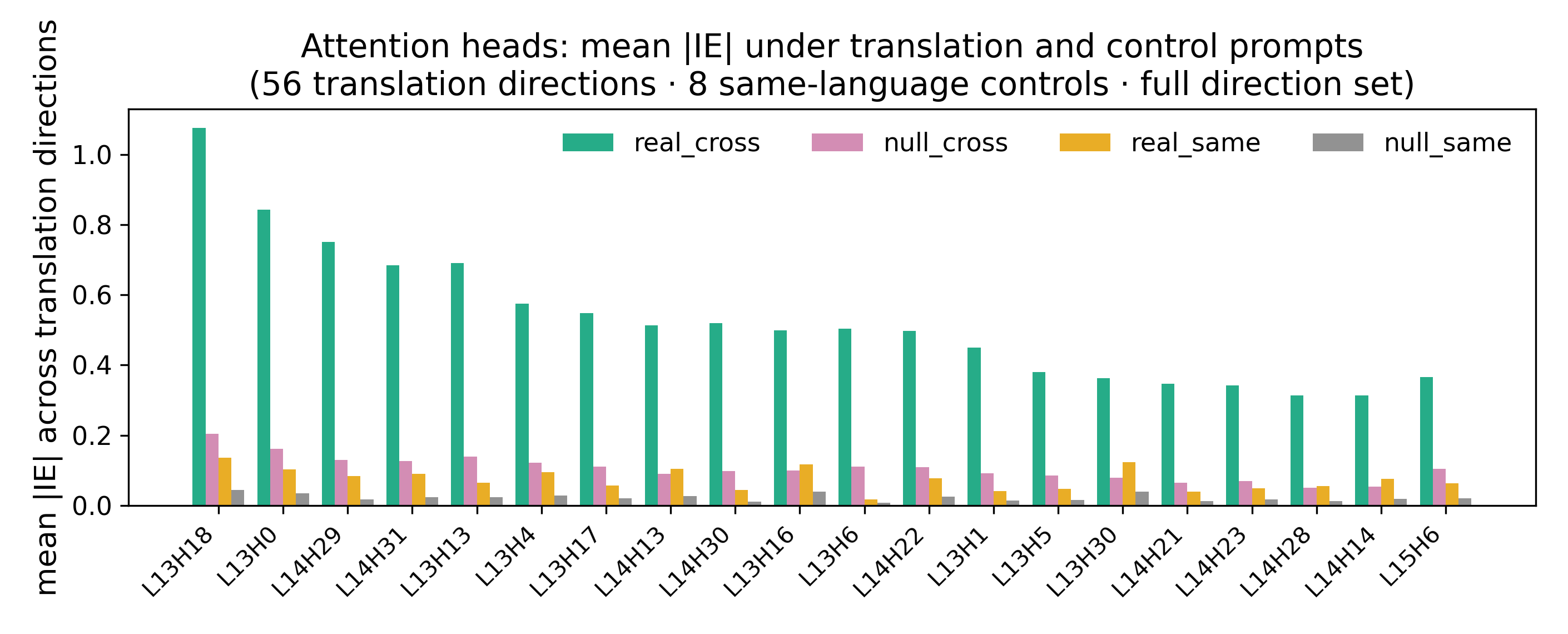}
    \caption{Llama-3.1-8B mean $|\widehat{\mathrm{IE}}|$ for top attention heads
    under translation and control prompts, averaged over 56 translation directions.
    The leading heads have much larger effects in the translation setting than in
    the controls, indicating that these heads are selective for matching translations and do not directly increase the probability of the given target sentence.}
    \label{fig:promptswap-llama-head-decomp}
    \end{figure*}
    
    We first search for model components that are causally relevant in promoting the valid translation over an invalid translation. We do so using generative causal mediation (GCM; \citealp{sankaranarayanan2026activationsteeringgenerativecausal}). Intuitively, GCM measures how intervening on the activation of a model component (e.g., an attention head) influences the model's relative preference for one continuation over another.
    
    We measure the model's preference for a single gold translation by swapping the source sentence that the model reads. Let $r$ be a gold translation of a source sentence, held fixed throughout. Let $p_{\text{orig}}$ be a prompt whose final source sentence is the one that $r$ actually translates, and let $p_{\text{cf}}$ be the same prompt with a distinct final source sentence. We define the preference metric $M$:
    \begin{equation}
    M = \log \pi(r\mid p_{\text{orig}}) - \log \pi(r\mid p_{\text{cf}}),
    \label{eq:metric}
    \end{equation}
    where $\log\pi(r\mid p) = \sum_t \log p(r_t \mid p, r_{<t})$ is the
    teacher-forced log-probability of a continuation, summed over its tokens. $M$ is the degree to which the model prefers to generate the gold translation $r$ when it has read the matching source, rather than a mismatched one; it isolates how much of the production of $r$ depends on having read the source content.
    
    We wish to know how strongly each attention head influences this preference. To measure this, we take the activation of an attention head; we call this $z$, and let $z_{\text{orig}}$ and $z_{\text{cf}}$ be the values it takes when the model reads $p_{\text{orig}}$ and $p_{\text{cf}}$ respectively. Let $\text{IE}(z)$ be the causal contribution of $z$ to $M$, measured as the change in $M$ when we move the head from its mismatched-source state $z_{\text{cf}}$ to its matched-source state $z_{\text{orig}}$ while holding the rest of the computation fixed. We source the counterfactual activation $z_{\text{cf}}$ by running a forward pass on $p_{\text{cf}}$ and caching $z$ at the final source-token position, then patch it into the corresponding position of the run we score:
    \begin{equation}
    \mathrm{IE}(z) = M(z_{\text{orig}}) - M(z_{\text{cf}}).
    \label{eq:patch}
    \end{equation}
    
    Patching a single head and rerunning the model tells us that head's exact contribution, but computing $\text{IE}(z)$ this way for every $z$ is intractable: it would require $O(Z\cdot n)$ forward passes, where $Z$ is the number of mediators and $n$ the number of examples. We instead use \emph{attribution patching} \citep{syed2023attributionpatchingoutperformsautomated}, a first-order linear approximation of the IE based on gradient attributions \citep{simonyan2013deep}:
    \begin{equation}
        \widehat{\mathrm{IE}}(z) = \nabla_z M\big|_{z=z_{\text{orig}}} \cdot (z_{\text{orig}} - z_{\text{cf}}).
        \label{eq:ie}
    \end{equation}
    The gradient $\nabla_z M$ for each component can be computed in a single backward pass, and the deltas $(z_{\text{orig}} - z_{\text{cf}})$ for each component can be computed in two forward passes by caching each component's activation on the two source prompts.
    
    The magnitude of the indirect effect $\widehat{\mathrm{IE}}$ is how much of a model's output behavior (as quantified by $M$) flows through a component when all else is kept constant. Because $M$ rewards producing $r$ under the correct source, components with a positive indirect effect are those that cause the probability of the correct translation $r_\text{orig}$ to increase relative to the counterfactual translation $r_\text{cf}$, and those with a negative indirect effect cause the probability of the correct translation to \emph{decrease} relative to the counterfactual translation.
    
    We instantiate $r$ as a gold translation from FLORES \citep{goyal-etal-2022-flores}, a massively parallel dataset that supports 101 languages. Given a source and target language, we construct a 2-shot prompt as follows:
    \begin{quote}\small\ttfamily
    
    \{Source\}: \{shot 1 source\}\\
    \{Target\}: \{shot 1 target\}\\
    \{Source\}: \{shot 2 source\}\\
    \{Target\}: \{shot 2 target\}\\
    \{Source\}: \{query source\}\\
    \{Target\}:
    \end{quote}
    and patch at the last source-token position (the final \texttt{": "}). For each of the 56 ordered pairs where the source and target can be any language in \{English, Spanish, German, French, Turkish, Arabic, Hindi, Hebrew\} (excluding pairs where the source and target are the same), we compute the $\widehat{\mathrm{IE}}$ for all attention heads
    over $n=100$ uniformly sampled pairs.
    
    Naïvely, we may find heads that respond at least in part to variations in the source samples that are unrelated to the translation task. To verify that the components we find are selective for correct source--target pairings, we compare indirect effects with three controls. First, the \textbf{same-language control} refers to cases where the source language is the same as the target, such that the correct ``translation'' is the same sentence copied from the source, and the counterfactual translation is a randomly sampled sentence in the same language as the source. Higher indirect effects for the translation setting than the same-language control indicate that the component specifically causes correct translations to be more probable, and not copies of the source sequence. The \textbf{null cross-language control} refers to a setup similar to the machine translation setup, but where \emph{neither} the original nor counterfactual completion are the correct translation. We expect indirect effects here to be very small relative to the translation task. Finally, we have the \textbf{null same-language control}, where both the original and counterfactual completions are in the same language as the source, but neither are the same as the source sentence.

    The heads we find have larger effects in the translation setting by far than in any control setting (Figure~\ref{fig:promptswap-llama-head-decomp}). The mean $|\widehat{\mathrm{IE}}|$ in the translation task exceeds that of the null cross-language control by $3.0\times$ over all heads, and $5.2\times$ over the top-10 heads for Llama (and $2.5\times$ for all heads/$4.3\times$ for the top-10 heads for Aya). We observe a similar magnitude of increase relative to both same-language controls. This suggests that the heads we have found are responsible for performing machine translation, and that their effects cannot be explained by their general utility in generating fluent target sequences regardless of the source.
    
    The translation-specific heads are localized to layers 13--14 in Llama and 15--20 in Aya. With respect to the stages of inference hypothesis \citep{lad2024the}, this would correspond to the layers that refine concept representations into increasingly abstract representations.
    
    \subsection{Translation Heads Have Similar Effects Across Language Pairs}
    
    An interlingua should have multilingual feature representations; this would allow a model to reuse the same representations when reading a source sequence and generating the target sequence. Prior work has established the existence of massively multilingual representations \citep{wendler-etal-2024-llamas}, including grammatical concept representations \citep{brinkmann-etal-2025-large}; here, we confirm these findings for the model we study by investigating their indirect effects across language pairs.
    
    We follow the procedure of \S\ref{sec:gcm} to compute $\widehat{\mathrm{IE}}$ for each language pair in the translation test dataset. We show the indirect effects for the top heads by absolute $\widehat{\mathrm{IE}}$ across language pairs; if there is feature reuse across pairs, then the sign of the effect should be the same for many language pairs. Note that the magnitude of $\widehat{\mathrm{IE}}$ is not directly comparable across languages, as the initial probability of the target sequence and its translation capabilities are language pair--dependent.
    
    \begin{figure}[t]
    \includegraphics[width=\linewidth]{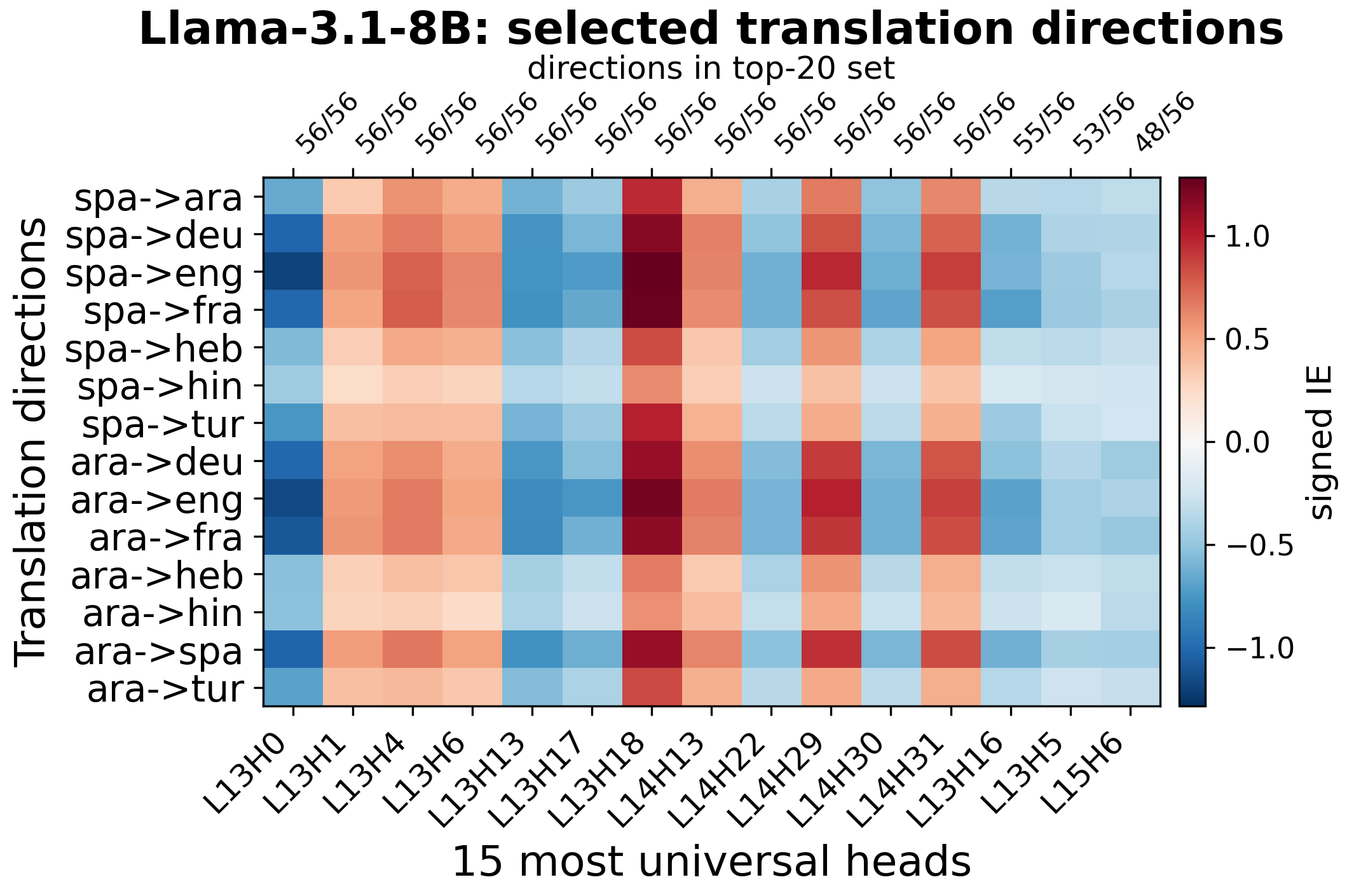}
    \caption{Signed mean IE for Llama-3.1-8B on machine translation for selected language pairs.
    Each column is 1 of the 15 top attention heads by $\widehat{\mathrm{IE}}$ across all languages. The top axis shows for how many language pairs the head was in the top-20 set. Several heads have stable effect signs across language pairs, which suggests that the same heads are being reused across many language pairs.}
    \label{fig:promptswap-llama-universal-heads-selected}
    \end{figure}
    
    The top attention heads by $\widehat{\mathrm{IE}}$ are largely shared across language pairs. Figure~\ref{fig:promptswap-llama-universal-heads-selected} shows the 15 heads that appear most often in a single direction's top-20 by $|\widehat{\mathrm{IE}}|$: the most universal of these are in the top-20 for all 56 directions. The sign of their effect is also stable---a given head keeps the same sign (favoring either correct or incorrect translations) across virtually every direction, regardless of source and target languages (For the full results from both models, see Figures ~\ref{fig:promptswap-llama-universal-heads-all} and ~\ref{fig:promptswap-aya-universal-heads-all}). 
    As predicted by our hypothesis, many of the translation mechanisms employed by the model do not depend on the choice of language pair; in fact, most of the top heads have the same directionality and general magnitude of effect on model performance for \emph{all} language pairs.
    
    \subsection{Ablating Translation Heads Degrades Performance on Translation and Monolingual Tasks}
    
    Another mechanistic prediction of our hypothesis is that the heads most responsible for performing machine translation should reuse computational machinery used for monolingual tasks. To test this, we first verify that ablating the top heads by indirect effect harms machine translation performance. Then, we show that ablating the same heads also harms performance on acceptability judgments and multiple-choice question answering, suggesting that these heads are not selective for translation; rather, they may be reusing computational machinery from more general language modeling mechanisms.

    \begin{figure}[t]
    \centering
    \includegraphics[width=\linewidth]{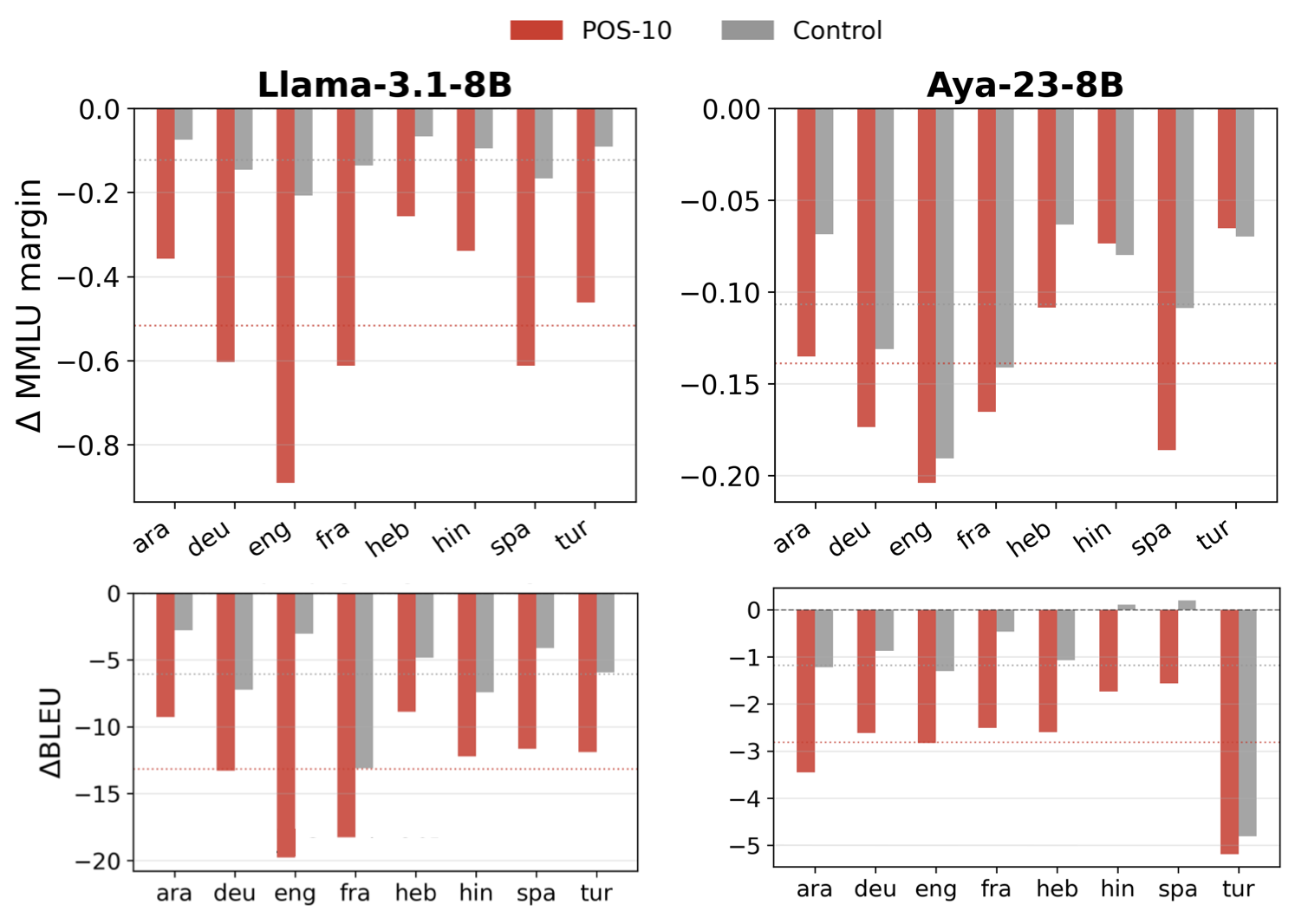}
    \caption{Ablation experiments overview. Ablating the top heads by causal influence on MT (POS-10, in red) performance drives a significant decrease in BLEU, and more than ablating random heads (Control, in grey). For GlobalMMLU, ablating those top heads reduces the model's ability to distinguish between the correct and incorrect answer more than ablating random heads does. The Control involves ablating the same number of heads in the same layers.}
    \label{fig:promptswap-ablation-overview-with-neg}
    \end{figure}
    
    For each target language $T$, we aggregate signed IE over the 7 language pairs in which $T$ is the target, then select the top 10 positive-IE heads (POS-10, the correct translation--favoring set) by signed magnitude. We mean-ablate a head by replacing its output with its average activation over the translation prompts at all token positions. As a control, we ablate the same number of randomly sampled heads from the same layers. We then regenerate FLORES translations and measure the change in BLEU relative to the original model before ablations.
    
    Ablating these heads causes significant reductions in BLEU. Ablating POS-10 degrades BLEU in all 8 target languages for both models (Figure~\ref{fig:promptswap-ablation-overview-with-neg}, bottom) by roughly twice the amount as the random control. This suggests that the heads we have found are causally relevant to translation performance. 

\begin{table*}[!h]
    \centering
    \small
    \setlength{\tabcolsep}{9pt}
    \begin{tabular}{lrrrrrr}
    \toprule
    & \multicolumn{3}{c}{Llama-3.1-8B}
    & \multicolumn{3}{c}{TinyAya-3B} \\
    \cmidrule(lr){2-4} \cmidrule(lr){5-7}
    Direction
    & Base & Mono & Parallel
    & Base & Mono & Parallel \\
    \midrule
    Xho\,$\rightarrow$\,Eng
    & 16.84 & 24.81 & 24.88
    & 23.46 & 26.64 & 27.80 \\

    \midrule
    
    Fra\,$\rightarrow$\,Eng
    & 43.27 & 42.28 & 38.87
    & 40.70 & 40.26 & 38.11 \\
    
    Deu\,$\rightarrow$\,Eng
    & 42.69 & 42.76 & 39.80
    & 40.20 & 41.21 & 39.58 \\
    \bottomrule
    \end{tabular}
    
  \caption{Monolingual fine-tuning recovers most of the Xhosa
translation gains obtained with parallel fine-tuning. Both conditions also
retain most performance on French and German translation (despite their not appearing in the fine-tuning data), although monolingual fine-tuning retrains slightly more performance.}
    \label{tab:mono-vs-parallel-bleu}
\end{table*}
    
    Having shown the importance of these heads for translation quality, we now ask whether these same heads are responsible for monolingual capabilities. For this, we again use the GlobalMMLU dataset. We uniformly sample 400 examples per language. Given prompt $p$ with correct token completion $r_\text{correct}$ and a randomly chosen incorrect completion $r_\text{incorrect}$, each minimal pair is scored by the acceptability margin
    \begin{equation}
    \Delta = \log p(r_{\text{correct}} \mid p) - \log p(r_{\text{incorrect}} \mid p)
    \end{equation}
    We report the \emph{change in} $\Delta$ after ablating the same heads ablated in the translation experiments:
    \begin{equation}
        \Delta_{\text{change}} = \Delta_{\text{ablation}} - \Delta_{\text{original}},
    \end{equation}
    where a negative $\Delta_{\text{change}}$ means the ablation weakened the model's ability to distinguish the correct from the incorrect answer.
    
    Ablating the POS-10 heads from the MT task results in a negative $\Delta_{\text{change}}$, whereas ablating random heads has a smaller effect (Figure~\ref{fig:promptswap-ablation-overview-with-neg}, top). That the same heads affect performance in both tasks provides preliminary evidence that these heads implement computations that are reused across tasks. Thus, these heads appear to implement computations that are not selective for translation alone. This provides further support for our hypothesis.\footnote{That said, applying the same ablations in MultiBLiMP results in a smaller effect; see Fig.~\ref{fig:ablate-multiblimp} in App.~\ref{app:mech}. Thus, these heads are not completely task-agnostic.}

    \section{Monolingual Fine-tuning Recovers Most Gains from Parallel Data}
    \label{sec:monolingual-vs-parallel}
    
    If LLMs translate via task-agnostic internal representations, translation quality should depend in part on the model's monolingual competence in the source and target languages. For example, if we wish to translate between a low-resource and high-resource language, then we should be able to improve translation performance given access only to data that improves language modeling quality on the low-resource language, assuming that we preserve capabilities in the high-resource language. Under this view, low translation performance may reflect a failure to encode the source sentence into an adequate internal representation, or to decode the target sentence fluently from it. Recent work has shown that bilingual or mixed-language signals during pretraining can be important for acquiring translation capabilities in LLMs \citep{briakou-etal-2023-searching,qorib-etal-2025-just,shao-etal-2026-role}. We ask the complementary question of whether, given a fully pretrained model, improvements in monolingual language modeling can (at least in part) transfer to translation even without parallel data. 

We test this idea using Llama-3.1-8B and TinyAya-3B.\footnote{We use TinyAya instead of Aya-23-8B because Aya-23-8B only provides an instruction-tuned model, and no base model. In pilot experiments, we found it difficult to achieve good performance after fine-tuning in any language (including high-resource languages) with instruction-tuned models.}
For each model, we compare how fine-tuning on an underrepresented language (Xhosa) affects translation performance when the training data consist of either parallel data or monolingual text.
Specifically, in the monolingual setting, we train on unaligned text consisting of  80\% Xhosa and 20\% English, French, and German data to avoid catastrophic forgetting of the model's existing language abilities.
In the parallel setting, we use paired Xhosa and English sentences from OPUS MT560 \citep{Gowda_2021}, presented in both translation directions.

Both settings use a next-token prediction loss and rank-16 LoRA adapters \citep{hu2022lora}. The reported runs use a learning rate of $3 \times 10^{-5}$, a maximum sequence length of 512 tokens, and one epoch over 100 million tokens. We evaluate translation with few-shot prompting on the FLORES devtest split~\citep{goyal-etal-2022-flores}, using examples from the dev split as in-context demonstrations. Thus, the models are prompted to translate at evaluation time even though the monolingual condition contains no translation examples during training.

    Table~\ref{tab:mono-vs-parallel-bleu} shows that monolingual fine-tuning recovers most of the improvement obtained with parallel data. 
    For Llama, monolingual fine-tuning raises BLEU from 16.84 to 24.81, compared with 24.88 under parallel fine-tuning. 
    This recovers 99\% of the improvement over the base model. 
    For Aya, monolingual fine-tuning raises BLEU from 23.46 to 26.64, compared with 27.80 under parallel fine-tuning, recovering 73\% of the improvement.
    For both models, both fine-tuning conditions largely preserve translation performance from French and German into English, although monolingual fine-tuning remains closer to the base performance.

These results are consistent with the view that once an LLM has been pretrained and acquired multilingual representations, improving its ability to model an underrepresented language can produce most of the available translation gains. 
Parallel data may still perform better because it can improve language modeling while also strengthening translation-specific mechanisms. 
Our results
therefore do not imply that parallel data are unimportant, particularly during pretraining. 
They show that, after pretraining, a large share of the gains from parallel fine-tuning can be achieved using only unaligned data.

We provide results for additional languages (German and Thai) and translation directions in Appendix~\ref{app:finetuning}. In short, changes in performance are smaller for higher-resource languages, but parallel and monolingual fine-tuning still achieve largely comparable results.

    \section{Discussion and Conclusions}
    We have provided evidence that machine translation capabilities in LLMs are mediated in significant part by multilingual representations that are also relevant for some monolingual tasks. 
    Across three complementary analyses, we find evidence consistent with this view: Translation performance is largely explained by source- and target-language competence, translation-relevant components overlap with components used for monolingual grammatical behavior, and monolingual Xhosa fine-tuning recovers most of the gains from parallel
fine-tuning across both models.
    Together, these results provide support for the hypothesis that LLMs translate by mapping source-language input into a latent feature space (that can also in theory be used for monolingual next-token predictions), and then generating target-language text from that shared representation.
    
    This interpretation helps explain why monolingual competence is strongly associated with translation performance. 
    If a model translates through a partially language-agnostic feature space, then improving its ability to read or write a language should improve translation involving that language. 
    The observation that target-language competence explains more variance than source-language competence may suggest that fluent and grammatical generation is often the bottleneck.
    Under this view, poor translation into a language need not imply the absence of a dedicated translation circuit for that language pair; instead, it may reflect weak target-language decoding capabilities given otherwise usable latent representations of the source sentence.

    We emphasize that translation-specific mechanisms not based on interlingua are also likely to exist; indeed, \citet{todd2024function} find that there are components selective for word translation. We do not claim that an interlingua is the only means by which LLMs translate, but rather, that it is a significant mechanism that controls a large proportion of model performance on MT tasks. It is likely that LLMs use a mixture of mechanisms \citep{gur-arieh2026mixing} to achieve their machine translation capabilities. 
    
    \section*{Limitations}
    The interlingua hypothesis is one plausible explanation for what we have observed, but our experiments do not rule out the existence of other mechanisms. The finding that the target language yields more predictive power than the source language in predicting BLEU score may be an artifact of the $n$-gram-based computation of BLEU scores. It is also possible that the features underlying high-quality translations are not well represented in the BLEU score.
    
    Our experiments were limited to two 8B-parameter models and one 3B-parameter model, and may not generalize to smaller or larger LLMs. Future work should investigate whether similar trends hold for a wider variety of language models, and whether recent developments in thinking models affect these findings.
    
    For our GCM experiments, we focus on components that mediate at the last token, missing computations that happen at earlier positions.
    
    Finally, our translation experiments are limited to a relatively small number of language pairs. Future work could scale up this experiment to investigate whether these trends hold across a large number of language pairs, and to what degree monolingual fine-tuning can recover the performance of parallel fine-tuning across many language pairs (and what other factors explain when this works well and when it does not).

    \section*{Acknowledgments}
    We are grateful to the members of the BAAIGL lab at Boston University for helpful comments on an earlier iteration of this work. The computational work reported on in this paper was performed largely on the Shared Computing Cluster, which is administered by Boston University’s Research Computing Services.
    Jannik Brinkmann is supported by the German Federal Ministry for Economic Affairs and the German Federal Ministry of Research, Technology and Space.
    
    \bibliography{custom}

    \appendix

    \section{Additional Results for Modeling Translation Performance}
    \label{app:modeling}

\begin{figure}[h]
    \centering
    \includegraphics[width=\linewidth]{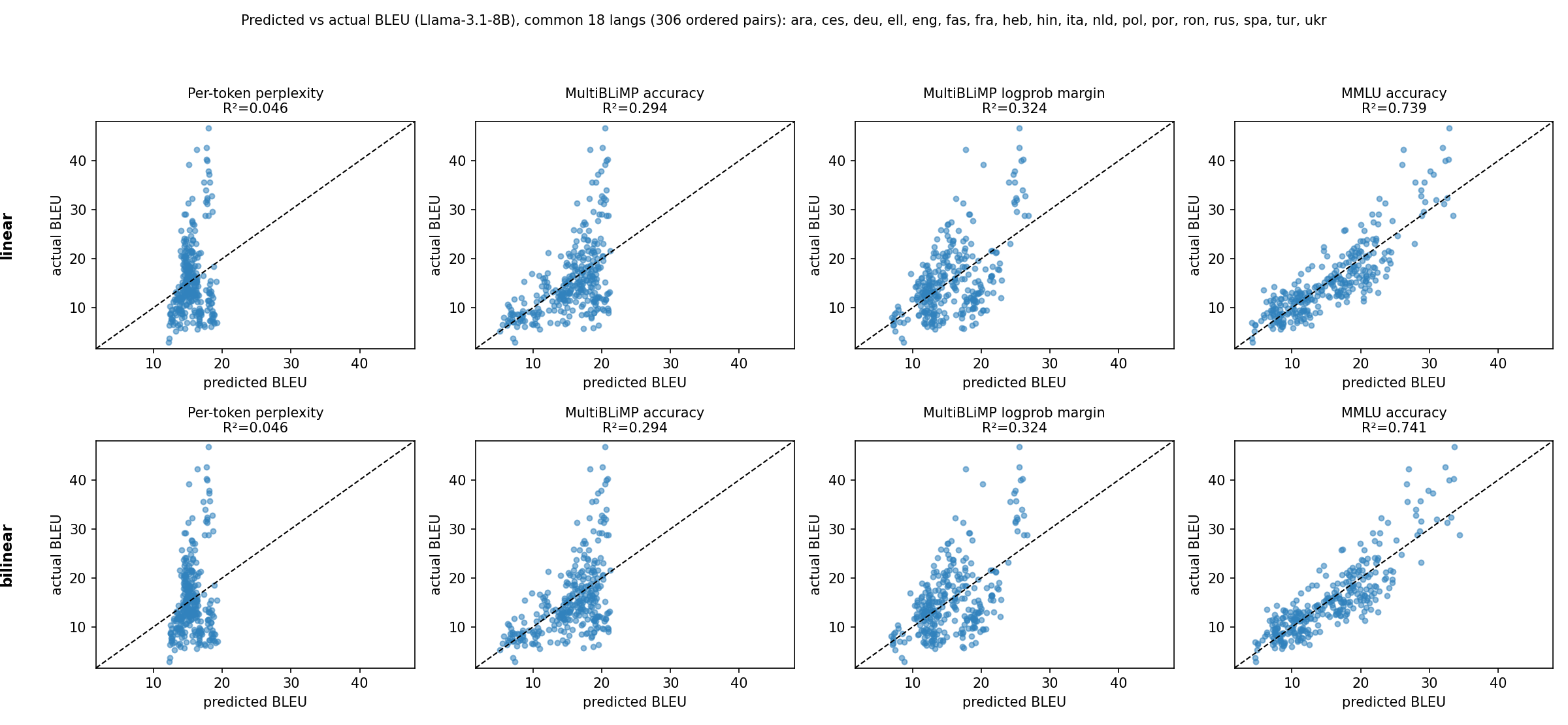}
    \includegraphics[width=\linewidth]{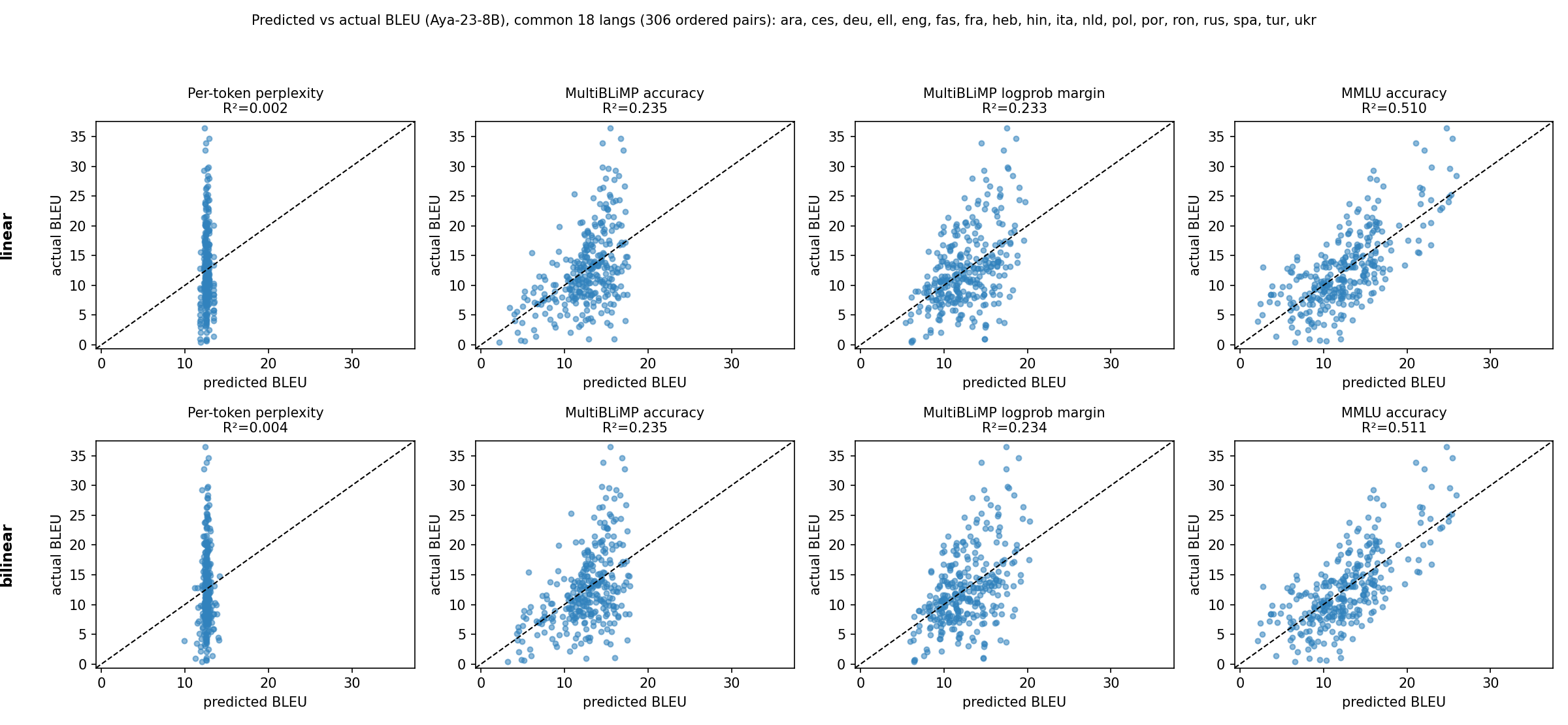}
    \caption{Predicted ($x$) vs.\ actual ($y$) BLEU on the common 18-language subset, for each proxy (columns) under the linear (top row) and bilinear (bottom row) models, Llama (top) and Aya (bottom). The two rows are visually indistinguishable for every proxy---the graphical form of the null interaction in Table~\ref{tab:proxy}.}
    \label{fig:grids}
    \end{figure}

    Figure~\ref{fig:grids} shows predicted vs.\ actual BLEU scores for the linear and bilinear models. In general, linear and bilinear models produce visually indistinguishable predictions.
    
    \label{app:grids}
     
    \subsection{Language-pair interaction tests}
    \label{app:interaction}
    Table~\ref{tab:interaction} reports the nested-model $F$-test on the $l_S\!\cdot\!l_T$ interaction coefficient, on the common 18-language subset and the 17-language no-English subset. The interaction is non-significant in every cell.
    \begin{table*}[t]
    \centering
    \small
    \setlength{\tabcolsep}{4pt}
    \begin{tabular}{llccc}
    \toprule
    Subset & Proxy (model) & $F(1,\cdot)$ & $p$ & $\Delta R^2$ \\
    \midrule
    18-lang & MMLU, Aya              & 0.001 & 0.980 & $<10^{-5}$ \\
    18-lang & MMLU, Llama            & 1.74  & 0.188 & 0.0015 \\
    18-lang & MultiBLiMP acc., Aya   & 0.262 & 0.609 & 0.0007 \\
    18-lang & MultiBLiMP acc., Llama & 0.000 & 0.990 & $<10^{-5}$ \\
    18-lang & MultiBLiMP margin, Aya & 0.400 & 0.528 & 0.0010 \\
    18-lang & MultiBLiMP margin, Llama& 0.028 & 0.867 & 0.0001 \\
    18-lang & Perplexity, Aya        & 0.439 & 0.508 & 0.0015 \\
    18-lang & Perplexity, Llama      & 0.009 & 0.924 & $<10^{-4}$ \\
    \midrule
    17-lang & MMLU, Aya              & 0.014 & 0.906 & $<10^{-4}$ \\
    17-lang & MMLU, Llama            & 0.955 & 0.329 & 0.0013 \\
    17-lang & MultiBLiMP acc., Aya   & 0.653 & 0.420 & 0.0019 \\
    17-lang & MultiBLiMP acc., Llama & 0.001 & 0.978 & $<10^{-5}$ \\
    17-lang & MultiBLiMP margin, Aya & 1.16  & 0.283 & 0.0038 \\
    17-lang & MultiBLiMP margin, Llama& 0.414 & 0.520 & 0.0014 \\
    17-lang & Perplexity, Aya        & 0.497 & 0.482 & 0.0018 \\
    17-lang & Perplexity, Llama      & 0.170 & 0.681 & 0.0006 \\
    \bottomrule
    \end{tabular}
    \caption{Nested $F$-test for the language-pair interaction term. Non-significant everywhere; largest $\Delta R^2=0.0038$.}
    \label{tab:interaction}
    \end{table*}
     
    \subsection{Predicting BLEU from Language-specific Translation Capabilities}
\begin{figure}[h]
    \centering
    \includegraphics[width=\linewidth]{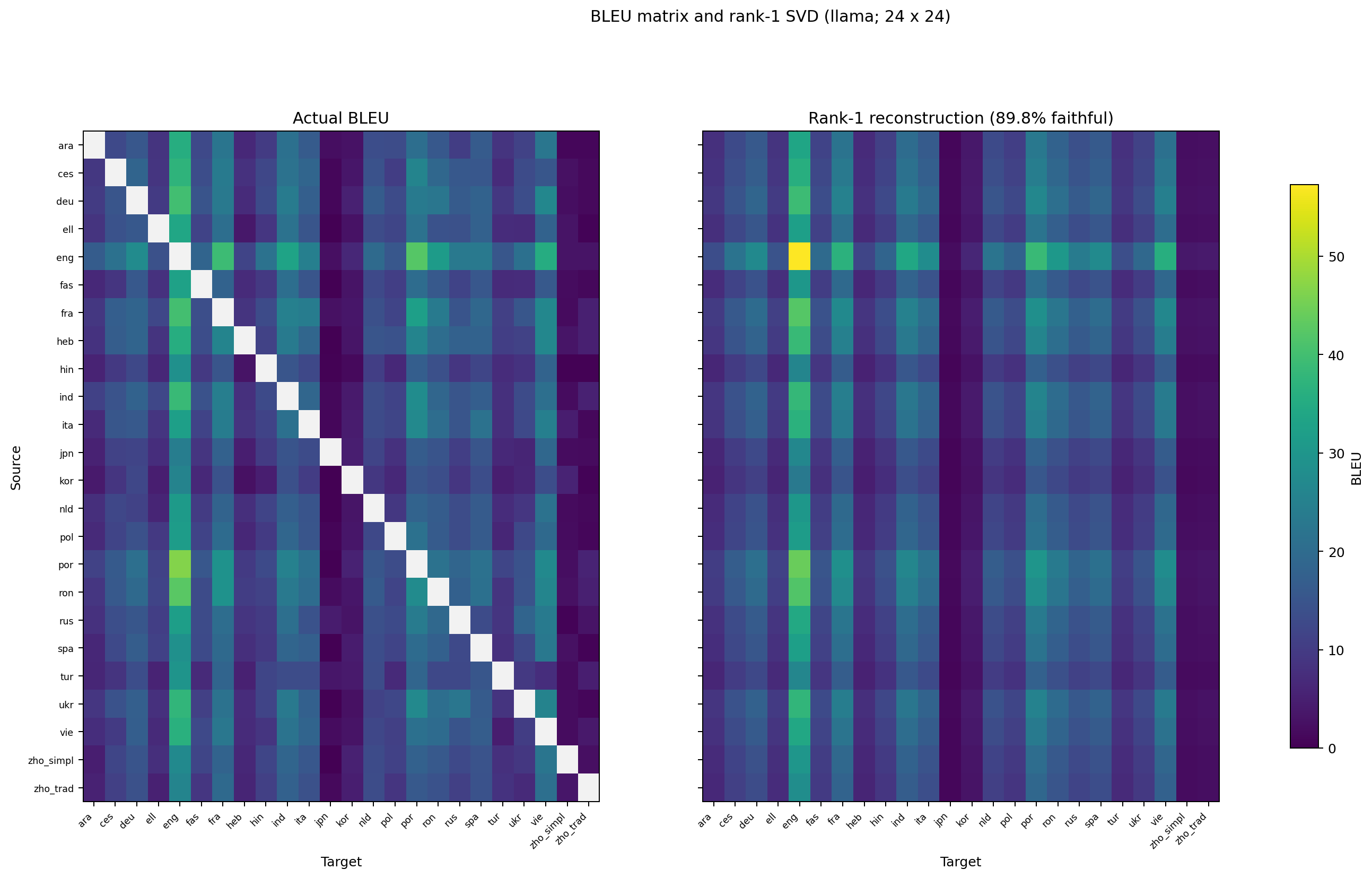}
    \caption{Left: the observed $24\times24$ Llama BLEU matrix (self-translation diagonal masked, never imputed). Right: its rank-1 reconstruction $u_S v_T$, fit by masked alternating least squares over the 552 observed off-diagonal cells. Faithfulness $=89.8\%$ (Aya: $83.4\%$); a single source-competence vector outer-producted with a single target-competence vector reconstructs most of the matrix.}
    \label{fig:rank1}
    \end{figure}

    Here, we show BLEU scores for all language pairs (Figure~\ref{fig:rank1}, left). We take each source or target language's average BLEU across language pairs, and fit linear models based on these terms to predict each language pair's BLEU score (see \S\ref{sec:linear_v_bilinear} for details). We observe that the error of a rank-1 linear predictor (Figure~\ref{fig:rank1}, right) is generally low at around 10--15\%. This provides further evidence that BLEU scores are predictable as a function of language-specific capabilities.
    
    \label{app:rank1}
     
    \subsection{Where the grammatical signal concentrates}
    \label{app:phenomena}
    Restricting the MultiBLiMP margin to subject--verb agreement phenomena \emph{improves} BLEU prediction (Llama, 18 languages: all-phenomena margin $R^2=0.324\to0.363$ for the SV-agreement subset). Per phenomenon, SV-Person reaches $R^2=0.688$, SV-Gender $0.519$, SV-Number $0.412$, versus subject--predicate agreement at $0.151$ / $0.037$. The production-side signal (agreement on the generated verb) is where translation predictiveness lives. We caution that per-phenomenon coverage is confounded with which languages each phenomenon is annotated for, so we report this as a strengthening analysis rather than a headline.
     
    \subsection{MMLU subject subsets}
    \label{app:mmlu-subsets}
    Unlike the MultiBLiMP phenomenon breakdown, the BLEU-predictive signal in MMLU is \emph{not} concentrated in any single subject category: aggregate MMLU ($R^2=0.598$ Llama / $0.499$ Aya, 23 languages) is stronger than every individual category (best single subset: Humanities $0.560$ for Llama, Social Sciences $0.413$ for Aya; STEM $0.360$/$0.190$). As with phenomena, subject-category accuracies are highly collinear across languages and largely track per-language resource level.
     
    \subsection{Functional form}
    \label{app:transforms}
    BLEU is bounded and right-skewed, and grammatical accuracy saturates near 1. Replacing the raw fit with $\log\text{BLEU}\sim\log(1-a_S)+\log(1-a_T)$ (a logit-like transform that un-saturates accuracy) improves $R^2$ from $0.250$ to $0.339$ (Llama) and $0.394$ to $0.444$ (Aya); the analogous log--log fit for perplexity improves $0.071\to0.127$ (Llama). We read these gains as variance-stabilization addressing the same ceiling effect that motivates the log-probability margin, rather than evidence of a specific (e.g.\ exponential) functional form, and so report the raw-scale linear models in the main text.

    \subsection{Does grammatical competence add signal beyond Global MMLU?}
    Adding MultiBLiMP source and target terms on top of the Global MMLU-only regression gives a small but significant gain (Table \ref{tab:nested}). Grammatical competence carries a little translation-relevant signal that general task accuracy doesn't capture.
    \begin{table*}[t]
    \centering
    \small
    \begin{tabular}{lcccc}
    \toprule
    Model & MMLU-only $R^2$ & Full $R^2$ & Partial $R^2$ & $F$-test $p$ \\
    \midrule
    Llama & 0.739 & 0.752 & 0.048 & $5.8\times10^{-4}$ \\
    Aya   & 0.510 & 0.556 & 0.094 & $3.7\times10^{-7}$ \\
    \bottomrule
    \end{tabular}
    \caption{Does grammatical competence add predictive value beyond general competence? We compare $\text{BLEU}\sim\text{MMLU}_S+\text{MMLU}_T$ to the same model alongside MultiBLiMP source and target terms. Adding MultiBLiMP accuracy adds some small signal for Llama and Aya.}
    \label{tab:nested}
    \end{table*}
     
    \section{Additional GCM Ablation Results}
    \label{app:mech}

    \subsection{Translation-specific head decompositions}
\begin{figure}[t]
    \centering
    \includegraphics[width=\linewidth]{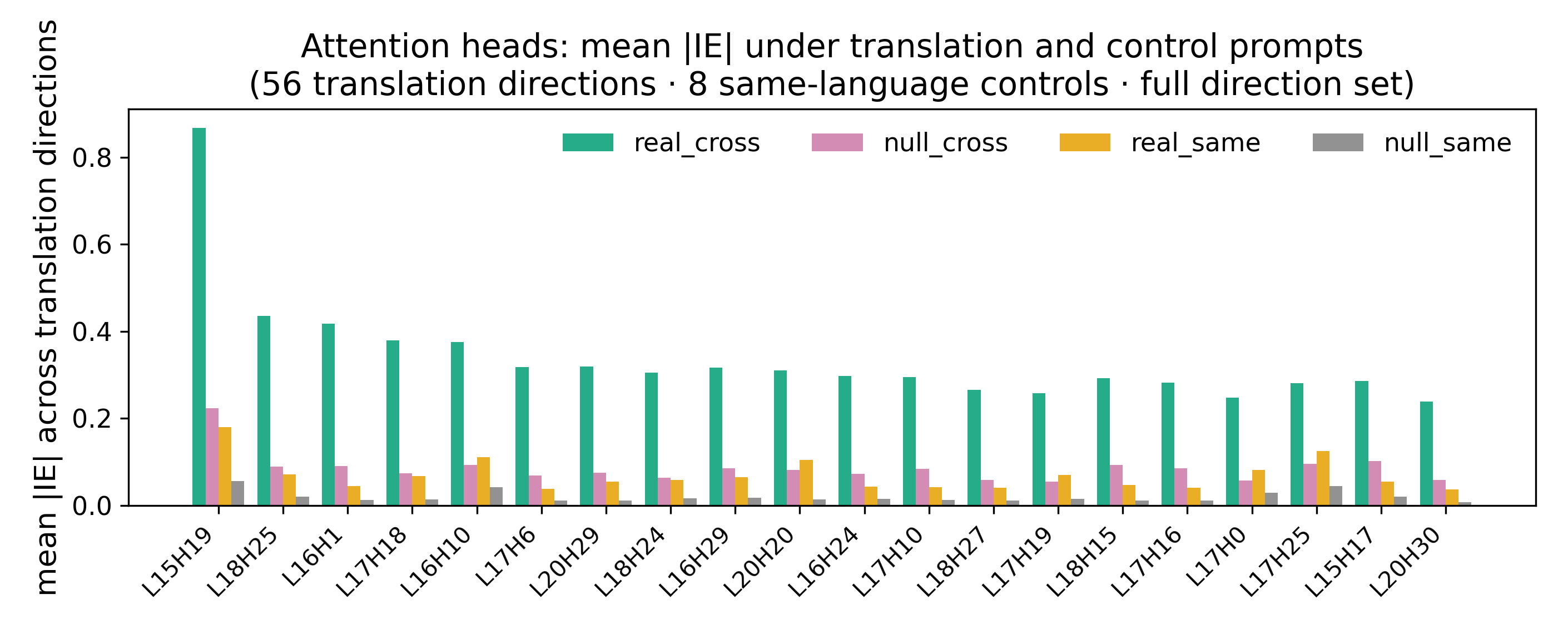}
    \caption{Prompt-swap GCM head IE under the four control tasks, top 20 heads
    sorted by $\textsc{real\_cross}-\textsc{null\_cross}$, for Aya-23-8B (cf.\
    Fig.~\ref{fig:promptswap-llama-head-decomp}). The top heads concentrate in layers 15--20.
    Support and bar definitions as in Fig.~\ref{fig:promptswap-llama-head-decomp}.}
    \label{fig:promptswap-decomp-aya}
    \end{figure}

    Figure \ref{fig:promptswap-decomp-aya} shows the attention head decompositions as in Fig.~\ref{fig:promptswap-llama-head-decomp} in the main text. The top heads concentrate in slightly later layers than Llama.
     
    \subsection{Top translation-specific heads}
    Tables \ref{tab:promptswap-heads} and \ref{tab:promptswap-heads-aya} show the details of the top 10 attention heads by translation-specific effect. A majority of the effect is concentrated in the top heads.
    \begin{table*}[t]
    \centering
    \small
    \setlength{\tabcolsep}{4pt}
    \begin{tabular}{rr rrrr r r}
    \toprule
    & & \multicolumn{4}{c}{mean $|\widehat{\mathrm{IE}}|$} & & \\
    \cmidrule(lr){3-6}
    Layer & Head & real\_cr & null\_cr & real\_sm & null\_sm & $\Delta$ & \#dir \\
    \midrule
    13 & 18 & 1.076 & 0.204 & 0.137 & 0.045 & 0.871 & 56 \\
    13 &  0 & 0.843 & 0.162 & 0.102 & 0.035 & 0.681 & 56 \\
    14 & 29 & 0.751 & 0.130 & 0.084 & 0.017 & 0.621 & 56 \\
    14 & 31 & 0.685 & 0.126 & 0.091 & 0.023 & 0.558 & 56 \\
    13 & 13 & 0.690 & 0.140 & 0.064 & 0.024 & 0.550 & 56 \\
    13 &  4 & 0.575 & 0.121 & 0.095 & 0.029 & 0.454 & 56 \\
    13 & 17 & 0.549 & 0.110 & 0.056 & 0.020 & 0.438 & 56 \\
    14 & 13 & 0.513 & 0.090 & 0.104 & 0.027 & 0.423 & 56 \\
    14 & 30 & 0.519 & 0.098 & 0.044 & 0.012 & 0.421 & 56 \\
    13 & 16 & 0.499 & 0.099 & 0.116 & 0.039 & 0.400 & 55 \\
    \bottomrule
    \end{tabular}
    \caption{Top 10 attention heads by translation-specific effect
    $\Delta=\textsc{real\_cross}-\textsc{null\_cross}$ under prompt-swap GCM (Llama), with mean $|\widehat{\mathrm{IE}}|$ in each of the four control
    tasks. \#dir is the number of the 56 cross-language directions in which the head
    ranks in that direction's top-30 by $\textsc{real\_cross}$ effect; all but one
    head appear in every direction.}
    \label{tab:promptswap-heads}
    \end{table*}
    \begin{table*}[t]
    \centering
    \small
    \setlength{\tabcolsep}{4pt}
    \begin{tabular}{rr rrrr r r}
    \toprule
    & & \multicolumn{4}{c}{mean $|\widehat{\mathrm{IE}}|$} & & \\
    \cmidrule(lr){3-6}
    Layer & Head & real\_cr & null\_cr & real\_sm & null\_sm & $\Delta$ & \#dir \\
    \midrule
    15 & 19 & 0.867 & 0.224 & 0.179 & 0.057 & 0.644 & 56 \\
    18 & 25 & 0.436 & 0.089 & 0.072 & 0.020 & 0.347 & 56 \\
    16 &  1 & 0.417 & 0.090 & 0.044 & 0.013 & 0.327 & 56 \\
    17 & 18 & 0.379 & 0.074 & 0.067 & 0.014 & 0.305 & 56 \\
    16 & 10 & 0.375 & 0.093 & 0.111 & 0.042 & 0.282 & 56 \\
    17 &  6 & 0.318 & 0.069 & 0.038 & 0.011 & 0.248 & 55 \\
    20 & 29 & 0.319 & 0.076 & 0.054 & 0.012 & 0.243 & 56 \\
    18 & 24 & 0.305 & 0.063 & 0.058 & 0.016 & 0.242 & 55 \\
    16 & 29 & 0.317 & 0.085 & 0.065 & 0.018 & 0.232 & 56 \\
    20 & 20 & 0.311 & 0.082 & 0.105 & 0.014 & 0.229 & 50 \\
    \bottomrule
    \end{tabular}
    \caption{Top 10 Aya attention heads by translation-specific effect
    $\Delta=\textsc{real\_cross}-\textsc{null\_cross}$ under prompt-swap GCM (cf.\
    Table~\ref{tab:promptswap-heads}). \#dir is the number of the 56 cross-language
    directions in which the head ranks top-30 by $\textsc{real\_cross}$ effect.}
    \label{tab:promptswap-heads-aya}
    \end{table*}
     
    \subsection{SAE-feature decomposition analyses}
\begin{figure*}[t]
    \centering
    \includegraphics[width=\linewidth]{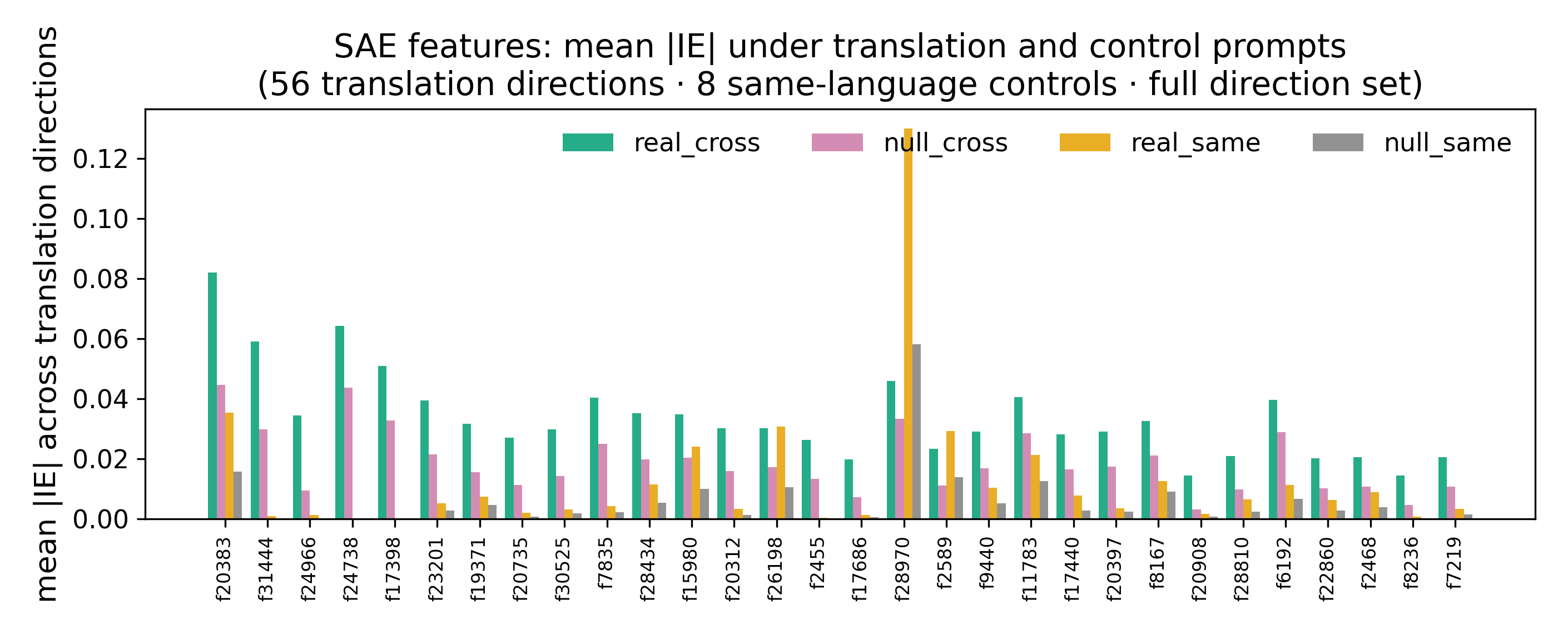}
    \caption{Llama-3.1-8B mean $|\widehat{\mathrm{IE}}|$ for top SAE features under
    translation and control prompts.}
    \label{fig:promptswap-llama-sae-decomp}
    \end{figure*}

\begin{figure*}[t]
    \centering
    \includegraphics[width=\linewidth]{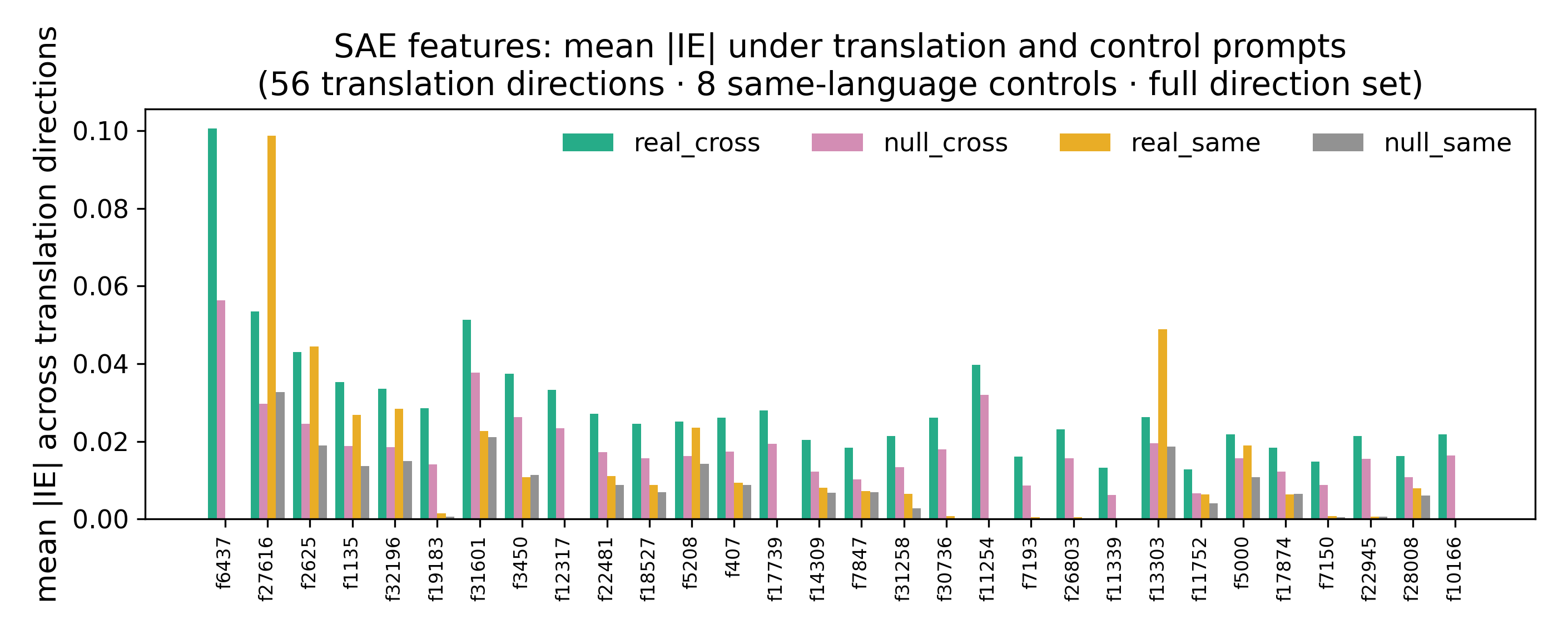}
    \caption{Aya-23-8B mean $|\widehat{\mathrm{IE}}|$ for top SAE features under
    translation and control prompts.}
    \label{fig:promptswap-aya-sae-decomp}
    \end{figure*}

    Top features show the same \textsc{real\_cross}-dominant pattern as attention heads but with weaker separation. A few top-ranked features are same-language features (Fig. $\ref{fig:promptswap-llama-sae-decomp}$ and $\ref{fig:promptswap-aya-sae-decomp}$).
     
    \subsection{Universality of translation-relevant heads}
\begin{figure*}[t]
    \centering
    \includegraphics[width=\linewidth]{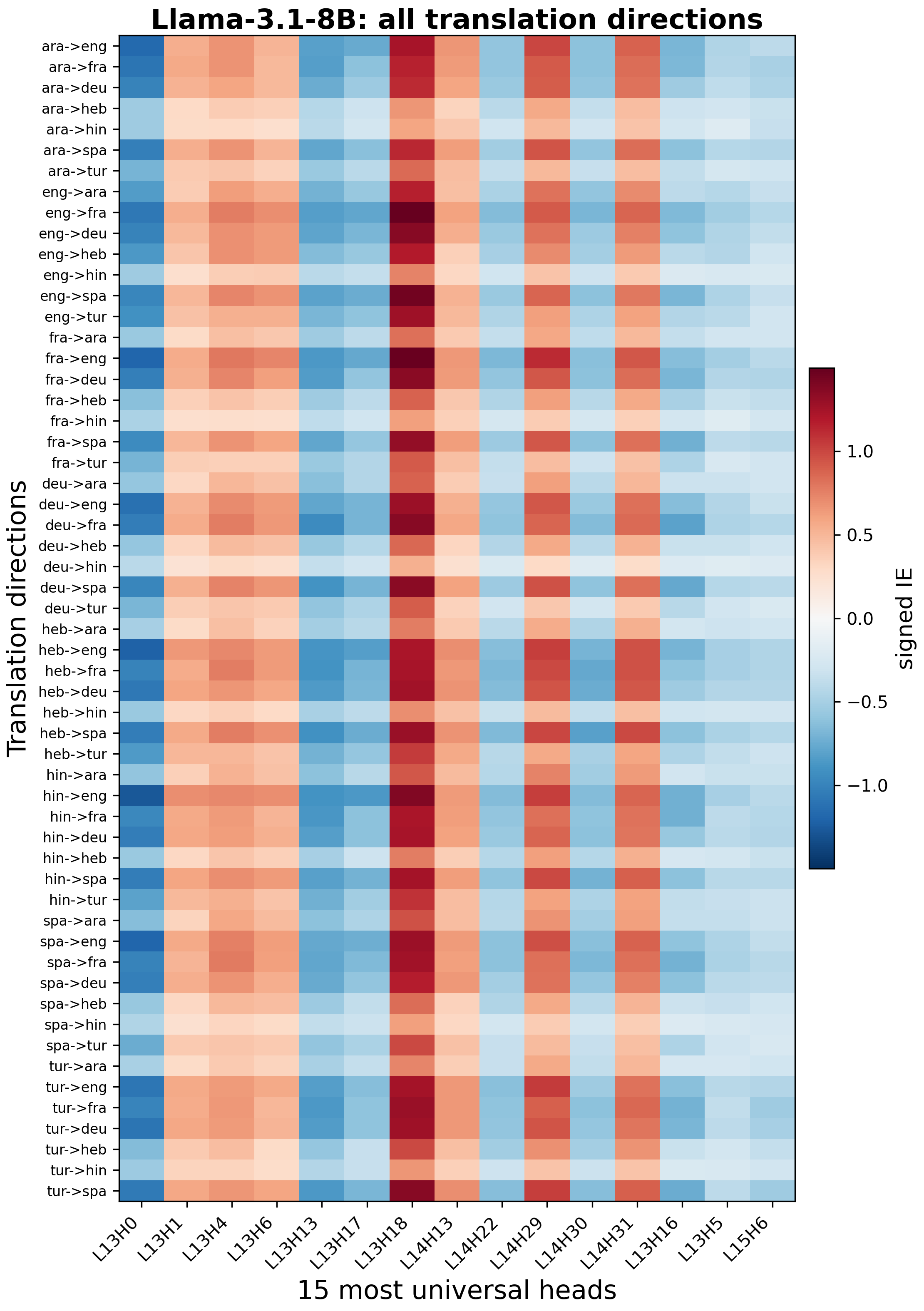}
    \caption{Full Llama-3.1-8B signed-IE heatmap for the 15 most universal
    prompt-swap heads across all 56 translation directions.}
    \label{fig:promptswap-llama-universal-heads-all}
    \end{figure*}

\begin{figure*}[t]
    \centering
    \includegraphics[width=\linewidth]{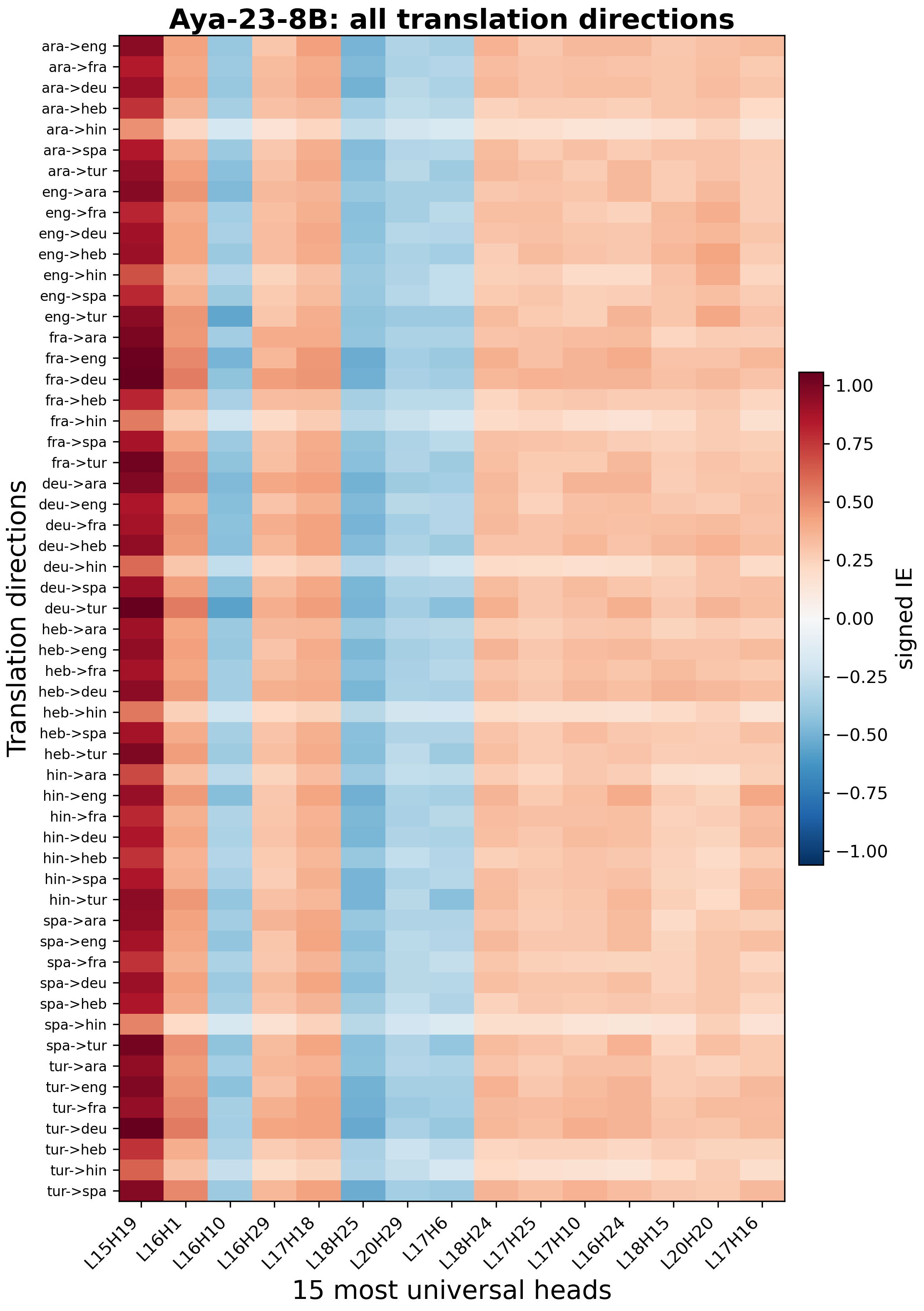}
    \caption{Full Aya-23-8B signed-IE heatmap for the 15 most universal prompt-swap
    heads across all 56 translation directions.}
    \label{fig:promptswap-aya-universal-heads-all}
    \end{figure*}

    Translation-relevant heads are universal across translation language pairs; Hebrew and Hindi show more divergence from the rest of the languages (Fig. \ref{fig:promptswap-llama-universal-heads-all} and \ref{fig:promptswap-aya-universal-heads-all}).

    \subsection{Head-IE concentration and sparsity}

\begin{figure}[t]
    \centering
    \includegraphics[width=0.8\linewidth]{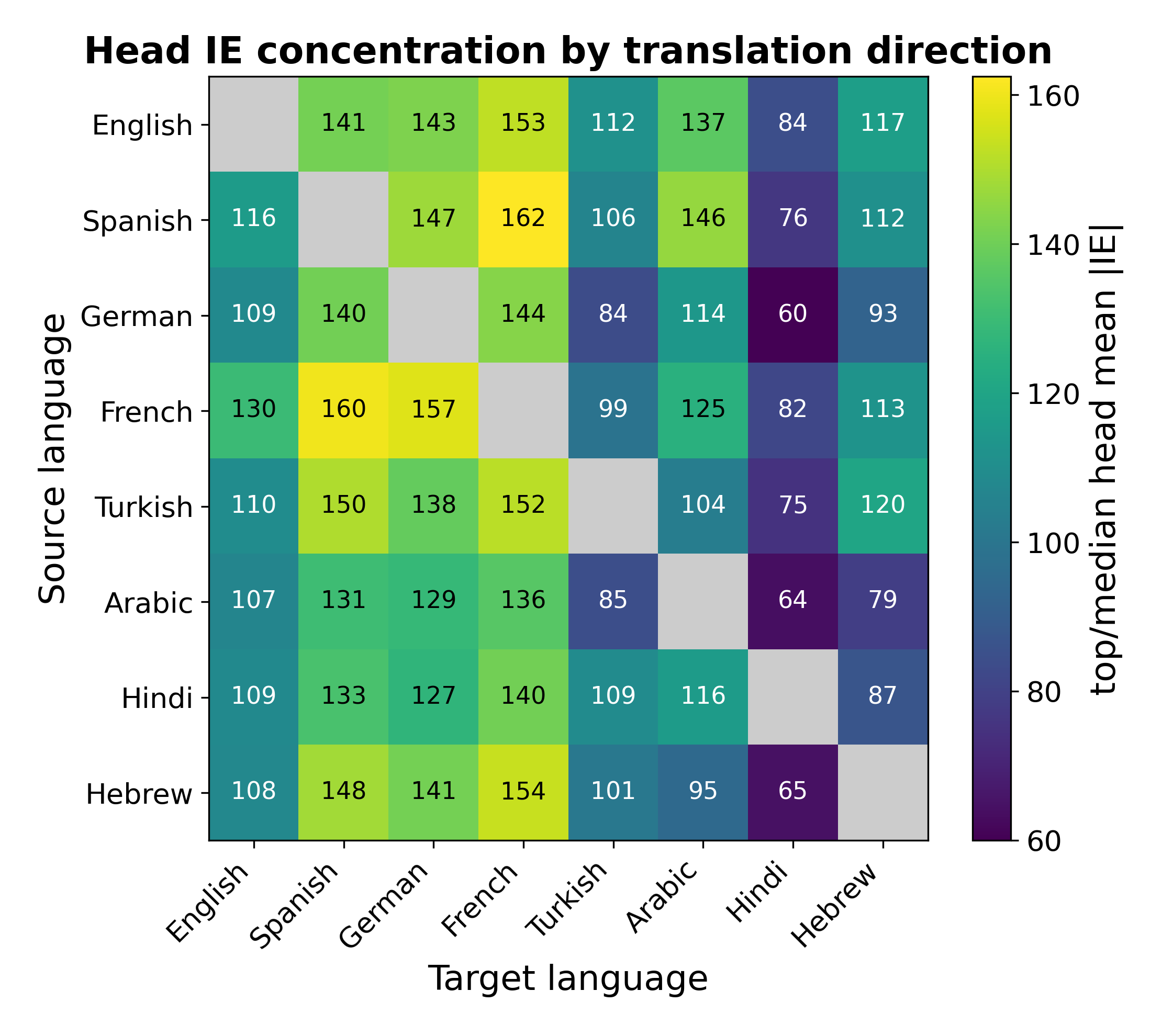}
    \caption{Per-direction head-IE sparsity under prompt-swap GCM: the ratio of the
    largest head's mean $|\widehat{\mathrm{IE}}|$ to the median head's, for each
    source$\to$target direction (Llama-3.1-8B; diagonal omitted). The ratio ranges
    from 60 to 162 (median 116). The top head is roughly two orders of magnitude
    above the median head in every direction, and is largest for directions into
    Spanish, German, and French and smallest into Hindi and Hebrew.}
    \label{fig:promptswap-sparsity}
    \end{figure}

\begin{figure}[t]
    \centering
    \includegraphics[width=0.8\linewidth]{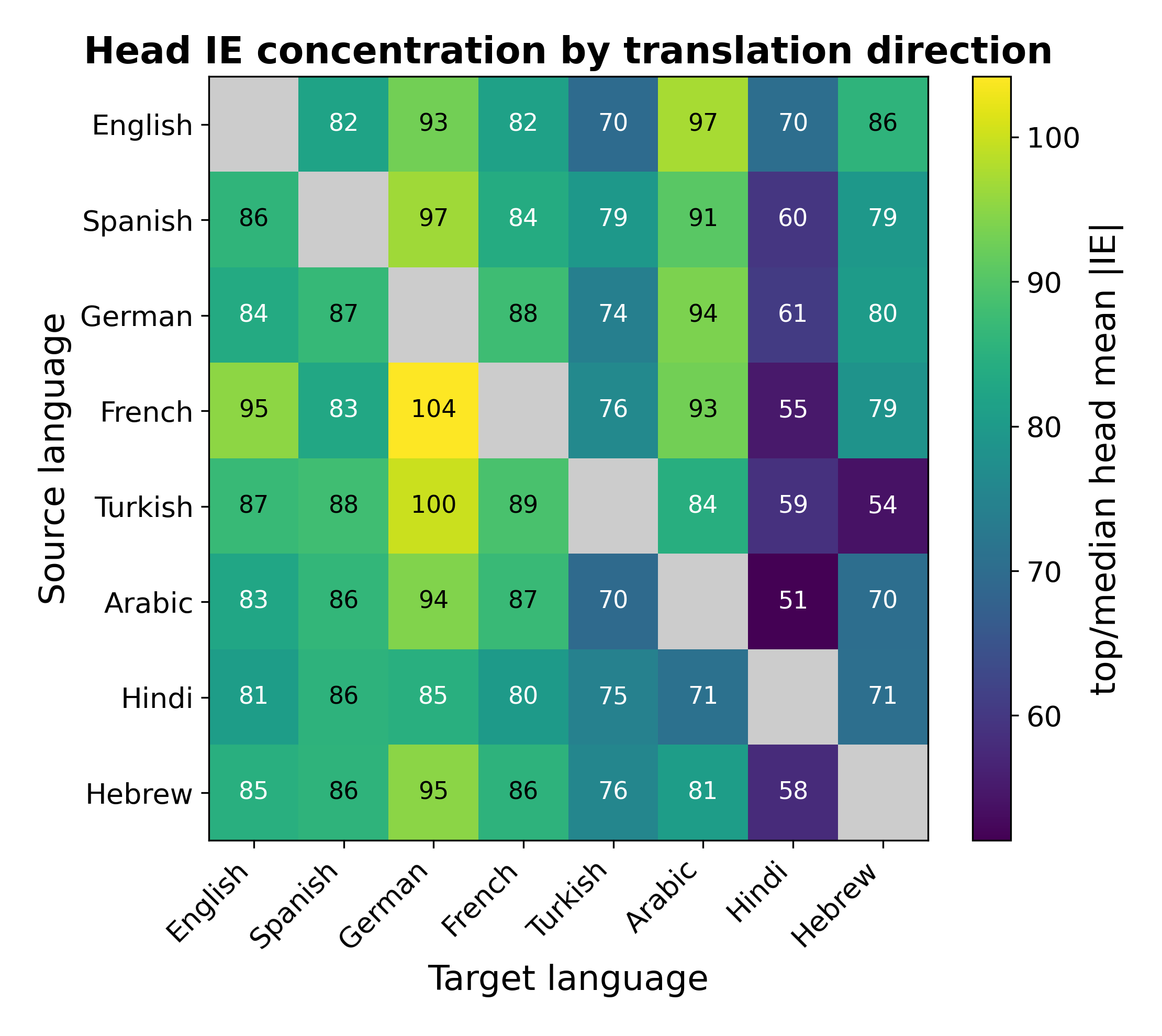}
    \caption{Per-direction head-IE sparsity for Aya-23-8B under prompt-swap GCM
    (cf.\ Fig.~\ref{fig:promptswap-sparsity}). The top-head/median-head ratio ranges
    from 51 to 104 (median 83).}
    \label{fig:promptswap-sparsity-aya}
    \end{figure}

    As further illustration of the head-IE sparsity, Fig. \ref{fig:promptswap-sparsity} and \ref{fig:promptswap-sparsity-aya} show that the ratio of the largest mean |IE| to the median ranges around 2 orders of magnitude.

\begin{figure}[t]
    \centering
    \includegraphics[width=\linewidth]{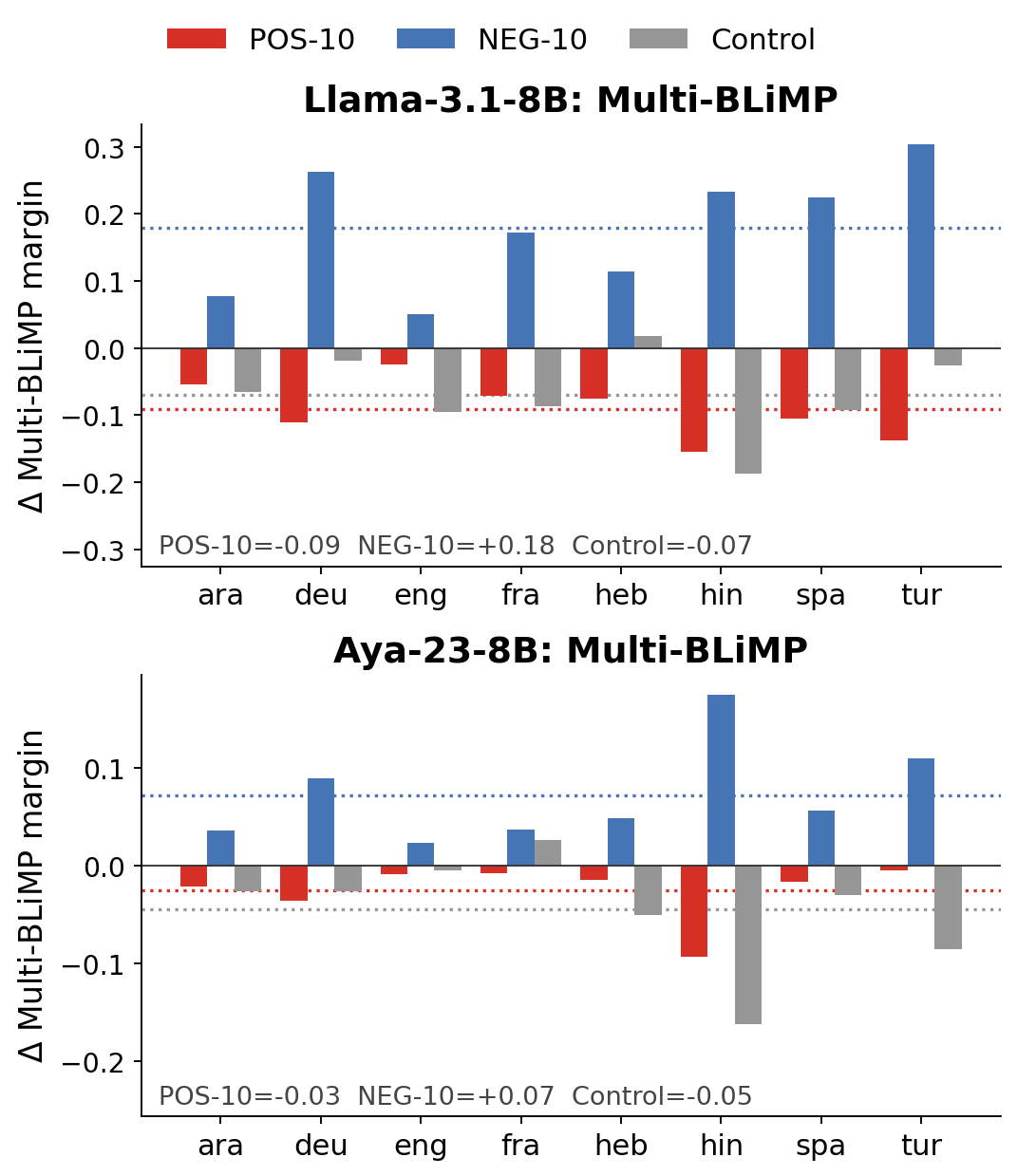}
    \caption{Change in the Multi-BLiMP grammaticality margin from mean-ablating the POS-10 and NEG-10 head sets and a random control. Dotted lines mark cross-language means.}
    \label{fig:ablate-multiblimp}
    \end{figure}

    The sign of the translation task selection carries over to Multi-BLiMP: ablating the POS-10 heads lowers the grammaticality margin in all 8 target languages for both models, and ablating the NEG-10 heads raises it in all 8 (Fig. \ref{fig:ablate-multiblimp}; Table \ref{tab:ablate-multiblimp}). The magnitude of the POS-10 effect, however, is comparable to that of ablating random heads from the same layers, so it is the consistent direction of the effects, rather than their size, that suggests these heads are reused across tasks.

\begin{figure}[t]
    \centering
    \includegraphics[width=\linewidth]{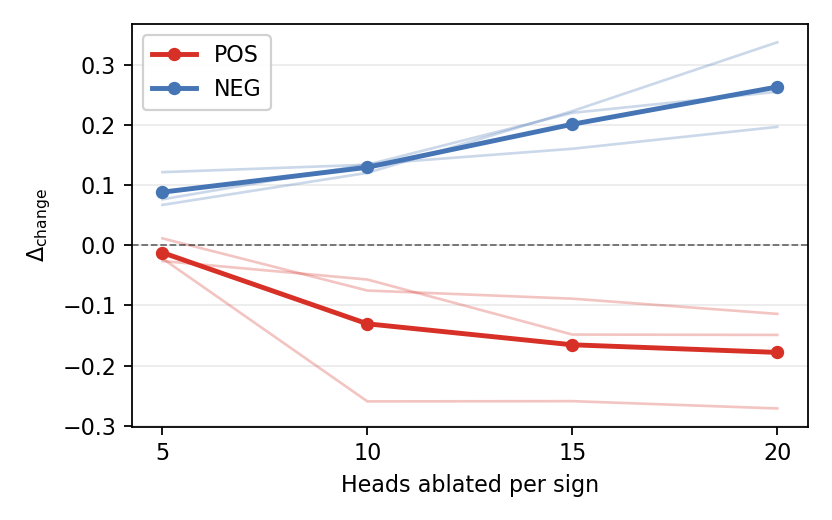}
    \caption{Head-count dose response (deu/eng/fra) on Llama. The effect is monotonic with respect to the number of heads ablated.}
    \label{fig:head-count-dose}
    \end{figure}

    \begin{table*}[t]
    \centering
    \small
    \setlength{\tabcolsep}{4pt}
    \begin{tabular}{l rrrrr rrrrr}
    \toprule
    & \multicolumn{5}{c}{Llama-3.1-8B} & \multicolumn{5}{c}{Aya-23-8B} \\
    \cmidrule(lr){2-6}\cmidrule(lr){7-11}
    target & base & POS-10 & ctrl$_{\mathrm{P}}$ & NEG-10 & ctrl$_{\mathrm{N}}$ & base & POS-10 & ctrl$_{\mathrm{P}}$ & NEG-10 & ctrl$_{\mathrm{N}}$ \\
    \midrule
    ara & 7.44 & $-0.055$ & $-0.066$ & $+0.077$ & $+0.050$ & 6.92 & $-0.021$ & $-0.026$ & $+0.035$ & $+0.009$ \\
    deu & 10.18 & $-0.111$ & $-0.019$ & $+0.263$ & $-0.076$ & 7.88 & $-0.037$ & $-0.026$ & $+0.089$ & $-0.023$ \\
    eng & 5.23 & $-0.025$ & $-0.095$ & $+0.050$ & $-0.002$ & 4.09 & $-0.009$ & $-0.006$ & $+0.023$ & $0.000$ \\
    fra & 9.16 & $-0.072$ & $-0.087$ & $+0.172$ & $+0.051$ & 6.33 & $-0.008$ & $+0.026$ & $+0.036$ & $0.000$ \\
    heb & 11.49 & $-0.075$ & $+0.018$ & $+0.113$ & $-0.108$ & 7.75 & $-0.015$ & $-0.050$ & $+0.049$ & $-0.013$ \\
    hin & 6.47 & $-0.155$ & $-0.188$ & $+0.232$ & $+0.012$ & 7.61 & $-0.094$ & $-0.162$ & $+0.175$ & $-0.077$ \\
    spa & 9.61 & $-0.105$ & $-0.093$ & $+0.224$ & $-0.069$ & 6.58 & $-0.017$ & $-0.031$ & $+0.056$ & $-0.015$ \\
    tur & 9.56 & $-0.138$ & $-0.027$ & $+0.304$ & $+0.093$ & 9.45 & $-0.005$ & $-0.085$ & $+0.109$ & $-0.071$ \\
    \midrule
    mean & 8.64 & $-0.092$ & $-0.069$ & $+0.179$ & $-0.006$ & 7.08 & $-0.026$ & $-0.045$ & $+0.072$ & $-0.024$ \\
    \bottomrule
    \end{tabular}
    \caption{Multi-BLiMP margin $\Delta_{\text{change}}$ relative to baseline from
    mean-ablating the POS-10 and NEG-10 GCM head sets and their size-matched controls (ctrl$_{\mathrm{P}}$ and ctrl$_{\mathrm{N}}$: 10 random heads drawn from the same layers as the corresponding set). Negative means ablation weakened the grammaticality preference. Ablating POS-10 lowers the margin in all 8 targets and ablating NEG-10 raises it in all 8, for both models; the size of the POS-10 effect, however, is comparable to that of its random same-layer control. $n=400$ pairs per language except heb ($n=200$) and hin ($n=100$).}

    \label{tab:ablate-multiblimp}
    \end{table*}
    The choice of the number of heads to ablate is unimportant to the overall effect; ablating different numbers of top heads has monotonic effect on performance (Fig. \ref{fig:head-count-dose}).

\clearpage

\section{Additional Fine-Tuning Results}
\label{app:finetuning}

This section reports the remaining fine-tuning results. We use the
same models, training conditions, and evaluation procedure described
in the main text.

\subsection{Additional directions involving Xhosa}

Table~\ref{tab:additional-xhosa-bleu} reports translation into Xhosa
from English, French, and German. Both fine-tuning conditions improve
performance across all three directions and both models. For Llama,
parallel fine-tuning performs best when translating from English and
French, while monolingual fine-tuning performs best when translating
from German. For Aya, monolingual fine-tuning performs slightly better
in all three directions. Thus, neither condition performs best in
every setting.

\begin{table}[!h]
\centering
\small
\setlength{\tabcolsep}{4pt}
\begin{tabular}{llrrr}
\toprule
Model & Method & Eng. & Fra. & Deu. \\
\midrule
Llama & Base        & 1.39 & 0.57 & 0.91 \\
Llama & Monolingual & 2.49 & 1.24 & \textbf{1.95} \\
Llama & Parallel    & \textbf{4.25} & \textbf{1.60} & 1.85 \\
\midrule
Aya & Base          & 2.47 & 1.42 & 0.96 \\
Aya & Monolingual   & \textbf{4.00} & \textbf{2.00} & \textbf{2.13} \\
Aya & Parallel      & 3.79 & 1.97 & 1.95 \\
\bottomrule
\end{tabular}

\caption{BLEU for translation into Xhosa. Column labels identify the
source language. Bold marks the best fine-tuned result for each model
and direction.}
\label{tab:additional-xhosa-bleu}
\end{table}

\subsection{Fine-tuning on higher-resource languages}

We also apply the same general pipeline to German and Thai using Llama-3.1-8B. 
Unlike Xhosa, both languages already have strong translation performance in the base model. As shown in Table~\ref{tab:high-resource-finetuning}, neither monolingual nor parallel fine-tuning produces a consistent improvement. 
Monolingual fine-tuning leaves performance largely unchanged, while parallel fine-tuning reduces performance in several directions.

One possible explanation is that German and Thai were already well represented during pretraining, leaving less room for improvement from continued training. 
Some of the fine-tuning documents may also have appeared in the pretraining corpus.

\begin{table}[!h]
\centering
\scriptsize
\setlength{\tabcolsep}{3pt}
\begin{tabular}{lrrrr}
\toprule
& \multicolumn{2}{c}{Adapted language}
& \multicolumn{2}{c}{Other directions} \\
\cmidrule(lr){2-3} \cmidrule(lr){4-5}
Method
& Into Eng. & From Eng.
& Fra. into Eng. & Xho. into Eng. \\
\midrule
\multicolumn{5}{l}{\textit{German adaptation}} \\
Base        & 42.69 & 34.94 & 43.27 & 16.84 \\
Monolingual & 43.48 & 34.55 & 43.16 & 19.13 \\
Parallel    & 42.95 & 33.26 & 42.53 & 14.90 \\
\midrule
\multicolumn{5}{l}{\textit{Thai adaptation}} \\
Base        & 32.60 & 15.21 & 43.27 & 16.84 \\
Monolingual & 32.19 & 15.19 & 43.34 & 16.75 \\
Parallel    & 30.98 & 10.31 & 42.46 & 14.95 \\
\bottomrule
\end{tabular}

\caption{BLEU after fine-tuning Llama-3.1-8B on German or Thai.
The first two result columns involve the adapted language; the final
two measure other translation directions.}
\label{tab:high-resource-finetuning}
\end{table}

\subsection{Preservation of monolingual abilities}

We also measure MultiBLiMP accuracy after fine-tuning Llama on Xhosa.
As shown in Table~\ref{tab:finetuning-multiblimp}, performance is
largely preserved. However, the base model is already near the maximum
score on all three languages. These results are therefore not strong
enough to determine whether fine-tuning causes cross-lingual transfer
of linguistic abilities.

\begin{table}[t]
\centering
\small
\setlength{\tabcolsep}{7pt}
\begin{tabular}{lrrr}
\toprule
Model & English & French & German \\
\midrule
Base & 99.35 & 98.70 & 97.91 \\
Xhosa monolingual & 99.48 & 98.43 & 98.17 \\
Xhosa parallel & 99.22 & 96.15 & 95.65 \\
\bottomrule
\end{tabular}

\caption{MultiBLiMP accuracy after fine-tuning Llama-3.1-8B.
Scores are percentages. The near-ceiling base scores make small
differences difficult to interpret.}
\label{tab:finetuning-multiblimp}
\end{table}

Only the monolingual condition includes replay data from
high-resource languages. Differences in retained performance therefore
cannot be attributed only to the use of monolingual rather than
parallel data. A parallel condition with similar replay data could
also reduce forgetting.

\section{Artifact Licenses}
    We use several existing scientific artifacts in our experiments. FLORES-101 \citep{goyal-etal-2022-flores} is released under a CC BY-SA 4.0 license. MultiBLiMP \citep{jumelet-etal-2026-multiblimp} is released under a CC BY 4.0 license. GlobalMMLU \citep{singh-etal-2025-global} is released under an Apache 2.0 license. For the Xhosa--English fine-tuning experiments, we use the English--Xhosa MT560 sentence-pair dataset \citep{Gowda_2021}; because OPUS MT560 aggregates data from multiple sources, downstream users should consult the original OPUS MT560 provenance information before redistributing or extending the dataset.
    
    We also use pretrained model checkpoints and evaluation software. Llama-3.1-8B is released under the Llama 3.1 Community License, and Aya-23-8B is released under CC BY-NC 4.0 with Cohere's acceptable-use addendum. SacreBLEU \citep{post-2018-call}, which we use to compute BLEU scores, is released under an Apache 2.0 license.
    
    In accordance with their licensing terms, all artifacts are used solely for research purposes.
    
    \section{Computational Budget}
    \label{app:compute}
    
    All experiments use 8B-parameter language models: Llama-3.1-8B and Aya-23-8B. We do not train any models from scratch. The compute cost is dominated by translation generation and scoring, activation caching, gradient-based attribution patching, ablation experiments, and LoRA fine-tuning for the Xhosa adaptation experiments.
    
    The total computational budget was approximately 200 GPU hours on NVIDIA A100/H100 GPUs. The largest individual runs were the causal mediation experiments, which require forward passes with activation caching and backward passes for attribution estimates across attention heads, and the LoRA fine-tuning sweeps over learning rate, LoRA rank, and number of training examples.
    
    For activation caching, we used \textsc{NNsight}~\citep{fiotto-kaufman2025nnsight}. 
    
    \end{document}